\documentclass{article}

\usepackage{microtype}
\usepackage{graphicx}
\usepackage{subfigure}
\usepackage{booktabs} % for professional tables
\usepackage{multirow,multicol}
\usepackage{enumitem}
\usepackage[dvipsnames]{xcolor}

\usepackage[colorlinks=true, allcolors=blue]{hyperref}
\usepackage{wrapfig}
\usepackage{amsmath,amsfonts,bm}

\def\eqref#1{equation~\ref{#1}}
\def\1{\bm{1}}

\def\vt{{\bm{t}}}

\def\vx{{\bm{x}}}
\def\vy{{\bm{y}}}
\def\vz{{\bm{z}}}
\def\vX{{\bm{X}}}
\def\vY{{\bm{Y}}}

\def\mtheta{{\bm{\theta}}}

\DeclareMathAlphabet{\mathsfit}{\encodingdefault}{\sfdefault}{m}{sl}
\SetMathAlphabet{\mathsfit}{bold}{\encodingdefault}{\sfdefault}{bx}{n}

\def\gD{{\mathcal{D}}}

\def\gN{{\mathcal{N}}}

\def\gX{{\mathcal{X}}}
\def\gY{{\mathcal{Y}}}
\def\gZ{{\mathcal{Z}}}

\usepackage[accepted]{icml2025}

\usepackage{amsmath}
\usepackage{amssymb}
\usepackage{mathtools}
\usepackage{amsthm}

\usepackage[capitalize,noabbrev]{cleveref}

\theoremstyle{plain}

\theoremstyle{definition}

\theoremstyle{remark}

\usepackage[textsize=tiny]{todonotes}

\icmltitlerunning{Does learning the right latent variables necessarily improve in-context learning?}

\begin{document}

\twocolumn[
\icmltitle{Does learning the right latent variables necessarily improve in-context learning?}



\icmlsetsymbol{equal}{$\dagger$}
\icmlsetsymbol{es}{*}

\begin{icmlauthorlist}
\icmlauthor{Sarthak Mittal}{equal,mila,udem}
\icmlauthor{Eric Elmoznino}{equal,mila,udem}
\icmlauthor{Leo Gagnon}{equal,mila,udem}
\icmlauthor{Sangnie Bhardwaj}{mila,udem,google}
\icmlauthor{Tom Marty}{mila,udem}
\icmlauthor{Guillaume Lajoie}{es,mila,udem}
\icmlauthor{Dhanya Sridhar}{es,mila,udem}
\end{icmlauthorlist}

\icmlaffiliation{mila}{Mila -- Quebec AI Institute}
\icmlaffiliation{udem}{Universit\'e de Montr\'eal}
\icmlaffiliation{google}{Google DeepMind}

\icmlcorrespondingauthor{Sarthak Mittal}{sarthmit@gmail.com}
\icmlcorrespondingauthor{Guillaume Lajoie}{guillaume.lajoie@mila.quebec}

\icmlkeywords{Machine Learning, ICML}

\vskip 0.3in
]



\printAffiliationsAndNotice{\icmlEqualContribution,\icmlEqualSupervision} % otherwise use the standard text.

\begin{abstract}
Large autoregressive models like Transformers can solve tasks through in-context learning (ICL) without learning new weights, suggesting avenues for efficiently solving new tasks. For many tasks, e.g., linear regression, the data factorizes: examples are independent given a task latent that generates the data, e.g., linear coefficients. While an optimal predictor leverages this factorization by inferring task latents, it is unclear if Transformers implicitly do so or instead exploit heuristics and statistical shortcuts through attention layers. In this paper, we systematically investigate the effect of explicitly inferring task latents by minimally modifying the Transformer architecture with a bottleneck to prevent shortcuts and incentivize structured solutions. We compare it against standard Transformers across various ICL tasks and find that contrary to intuition and recent works, there is little discernible difference between the two; biasing towards task-relevant latent variables does not lead to better out-of-distribution performance, in general. Curiously, we find that while the bottleneck effectively learns to extract latent task variables from context, downstream processing struggles to utilize them for robust prediction. Our study highlights the intrinsic limitations of Transformers in achieving structured ICL solutions that generalize, and shows that while inferring the right latents aids interpretability, it is not sufficient to alleviate this problem.
\end{abstract}

\begin{figure*}
    \centering
    \includegraphics[width=\textwidth]{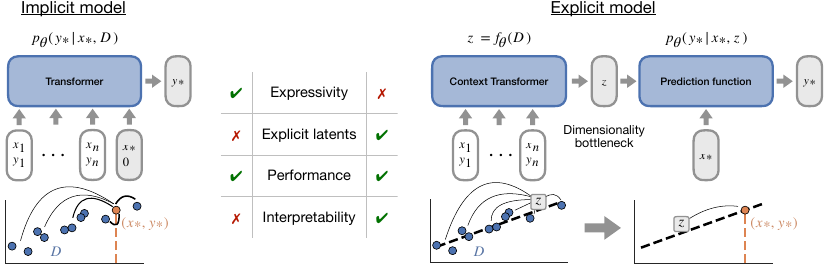}
    \vspace{-3mm}
    \caption{We compare the benefits of the implicit (\textit{left}) and the explicit (\textit{right}) model. Explicit models disentangle context aggregation and prediction into two separate functions with an inductive bias for inferring generative latent variables in order to solve the task. Implicit models are more expressive, but can learn non-parametric shortcut solutions that bypass latent variable inference.}
    \vspace{-6mm}
    \label{fig:explicit_implicit}
\end{figure*}
\vspace{-7mm}
\section{Introduction}
\vspace{-1mm}
\looseness=-1
% A notable ability of Transformer-based \citep{vaswani2017attention} large language models (LLMs) is their capacity to leverage demonstrations within their input to learn new tasks at inference time: in-context learning (ICL) \citep{bubeck2023sparks}. Although ICL underpins much of the capabilities of modern LLMs, including prompt engineering and chain-of-thought, its well documented brittleness is an active issue [\textcolor{red}{citations}]. Specifically, several papers \citep{Tang_2023,mccoy-etal-2019-right, mccoy2023embers} find that rather than extracting the tasks' underlying principles from the demonstrations, Transformers often rely on shortcuts directly comparing demonstrations to new datapoints, risking poor generalization.
% In this work, we investigate if biasing Transformers against these shortcuts can give to more systematic solutions which generalize better. Instead of studying ICL in LLMs, where lack fine-grained control over the data multiple related factor influence performance, we introduce controlled experimental settings that involve training Transformers from scratch on synthetic ICL tasks that are sufficiently complex, but where the relevant latent variables are well-understood.
Recent advancements in large language models \citep[LLMs,][]{radford2019language} showcase the Transformer architecture's \citep{vaswani2017attention} ability to perform novel tasks at inference through in-context learning \citep[ICL,][]{brown2020language}. Indeed, LLMs can learn from instructions and demonstrations provided as input, without requiring gradient-based learning. While ICL plays a key role in many LLM abilities \citep{lu2024emergentabilitieslargelanguage}, such as instruction-following \citep{wei2022finetunedlanguagemodelszeroshot} and chain-of-thought reasoning \citep{wei2023chainofthoughtpromptingelicitsreasoning}, the factors that influence its generalization -- particularly in out-of-distribution (OOD) settings -- remain poorly understood. Although ICL can leverage a combination of instructions and demonstrations, our analysis focuses specifically on its ability to model predictions based on a task’s training examples (demonstrations) provided in-context \citep{lampinen2024broader} beyond the modality of language.
% 
% Recent achievements of large language models \citep[LLMs;][]{radford2019language,brown2020language} showcase the Transformer architecture's~\citep{vaswani2017attention} ability to solve novel tasks at inference via in-context learning, the capacity to learn from instructions and demonstrations given as input without any gradient-based learning. Although ICL underpins many LLM capabilities, including prompt engineering \citep{...} and chain-of-thought \citep{...}, we do not yet understand which factors influence its ability to generalize, especially to out-of-distribution context and query sequences. While ICL can often describe task acquisition by LLMs via a combination of instructions and/or demonstrations, we limit our analysis to its ability of modeling predictions conditioned on a task's training set (demonstrations) passed in-context~\citep{lampinen2024broader}.

\looseness=-1
A plausible hypothesis behind the success of ICL is that since many tasks are based on some low-dimensional latents (e.g., complex games are described completely through their rules, linear regression through its underlying coefficients), Transformers might generalize to novel queries by inferring the \textit{task latents} from the context~\citep{hendel2023incontext,todd2024function,yang2025task}. This describes a \textit{parametric} \citep{bishop2006pattern} modeling approach that breaks the prediction mechanism into two parts: 1) inferring the task latents (i.e. parameters) from the context, and then 2) using them to make predictions on novel queries. With the right prediction function, such an approach leads to systematic OOD generalization to new queries.
%For instance\todo{Should we remove this sentence?}, in a linear regression task, a model could first try to infer the underlying weight vector used to generate the data, and subsequently use these inferred weights to make predictions for queries. 

\looseness=-1
However, mounting evidence \citep{wang2023label,han2023explaining,song2024shortcutlearningincontextlearning, tang-etal-2023-large, bhaskar2024heuristiccoreunderstandingsubnetwork,crosbie2024inductionheadsessentialmechanism} suggests that Transformers instead often employ more shallow solutions which rely on direct comparison of the query to demonstrations \citep[reminiscent of induction heads,][]{olsson2022context}. This is closer in spirit to \textit{non-parametric} approaches (such as nearest neighbors or kernel regression) which are known for their flexibility but poor generalization \citep{bishop2006pattern}. Since these solutions do not model the data generative process, they can be described as statistical shortcuts and risk poor performance on OOD context and queries -- e.g., learning the actual linear predictor for linear regression can generalize to any distribution over training and test points, but nearest-neighbour based interpolation might not. Interestingly, the functional form of attention operations is almost identical to that of kernel regression \citep{tsai2019transformer, han2023explaining}, making such solutions more natural for Transformers to express \citep{zhou2023algorithms}.

\looseness=-1
In this paper, we aim to test the hypothesis that biasing Transformers against non-parametric solutions can improve their ICL performance by encouraging parametric modeling. We minimally modify the Transformer architecture to prevent such non-parametric shortcuts and compare the OOD performance of the resulting model to that of a traditional Transformer on a large array of ICL tasks. We call this altered architecture an \textit{explicit model} because of its inductive bias of explicitly extracting structured latent variables to solve the tasks, and call the traditional Transformer architecture an \textit{implicit model}. Specifically, the explicit model prevents the query from directly attending to demonstrations in the context 
% but only to more global information aggregated from the context 
by introducing a bottleneck between the processing of the context and the query (see \autoref{fig:explicit_implicit}), similar to a conditional neural process \citep{garnelo2018conditional}.
To study the impacts of this inductive bias favoring parametric solutions, we need to establish that explicit models successfully recover task latents.
As such, we consider synthetic and real tasks for which the latent mechanisms are well understood, and systematically analyse the impact of task latents on generalization by comparing explicit and implicit models.

\looseness=-1
We find that the explicit model does not outperform the implicit one on OOD data, challenging the aforementioned hypothesis that avoiding non-parametric solutions enhances generalization. Our investigation into this lack of improvement reveals that the issue often lies in the explicit model's prediction function which has to leverage the inferred latent variables for downstream predictions on the query. Our controlled experiments and analysis on the interpretable nature of the bottleneck revealed strong evidence that while the explicit model often extracts relevant task latents, these are not properly utilized by the prediction function.

\looseness=-1
While on one hand, our research demonstrates that using a simple bottleneck in a Transformer can improve interpretability and explicitly extract task-relevant latent variables, it also suggests that the limitations of Transformers in learning more structured and generalizable ICL solutions are not solely due to non-parametric shortcuts that skirt latent variable inference, but due to more fundamental architectural limitations. 
In sum, our contributions are:
\begin{itemize}[noitemsep,topsep=0pt,leftmargin=3mm]
    \item Formalizing a framework to test whether parametric ICL solutions generalize better out-of-distribution.
    % Formalizing an experimental framework to evaluate the hypothesis that parametric ICL solutions generalize better out-of-distribution.
    \item Analyzing the benefits, or lack thereof, of inferring the true latents explicitly.
    \item Highlighting flaws in the prediction function and downstream latent utilization which hinders generalization.
    % \item Identifying shortcomings in the prediction function and the downstream utilization of learned latents in Transformers, which leads to poor generalization.
\end{itemize}

\begin{figure*}
    \centering
    \includegraphics[width=\linewidth]{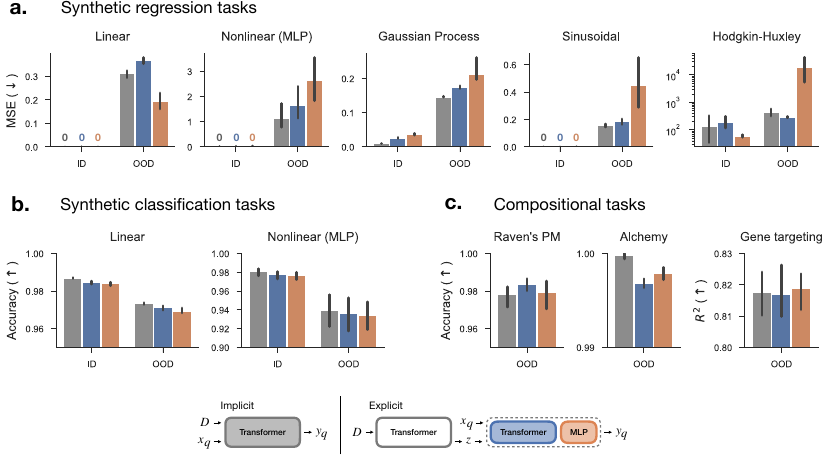}
    \vspace{-4mm}
    \caption{Comparison of implicit and explicit models in-distribution (ID) and out-of-distribution (OOD) across various domains: (a) synthetic regression, (b) classification, and (c) compositional generalization tasks. Implicit models are in shown \textcolor{gray}{gray}, explicit models with Transformer prediction in \textcolor{NavyBlue}{blue}, and with MLP prediction in \textcolor{Orange}{orange}. 
    Further details about tasks is provided in \autoref{section:task}.}
    \vspace{-6mm}
    \label{fig:results}
\end{figure*}
\vspace{-3mm}
\section{Notation}
\vspace{-1mm}
\looseness=-1
Throughout the paper, we denote datasets with the symbol $\gD$ which consists of a set of observations with inputs denoted via $\vx \in \gX$ and their corresponding outputs as $y \in \gY$. A task is defined by a functional mapping $g: \gX, \gZ \to \gY$ which maps observations $\vx$ to labels $y$ through some latents or parameters $\vz$, eg. $y = \vz^T\vx$ for a linear regression task, or $y \sim \gN(\cdot; \vz^T\vx, \sigma^2)$ for its stochastic counterpart. To ease readability, we will reserve $\vx_* \in \gX$ for the query, i.e. the test time observation we want to generalize to, and $y_* \in \gY$ its corresponding target. Finally, $\psi$ denotes the parameters of the context aggregation component of explicit model, which inputs the dataset $\gD$ and infers the corresponding parameters $\vz_\psi(\gD)$, and $\gamma$ the parameters of the prediction model which given a query $\vx_*$ and parameters $\vz \in \gZ$, provides the prediction. For the implicit model, these operations are subsumed into a single model, with parameters $\varphi$.

\vspace{-3mm}
\section{Implicit vs. Explicit Inference}
\vspace{-1mm}
\looseness=-1
We look at ICL in the context of algorithmic problems where the task is to predict the target $y_*$ from a query $\vx_*$ when provided with some context examples $\gD = \{(\vx_i, y_i)\}_{i=1}^n$, sharing a common underlying structure defined by the task latent $\vz$ and a functional form $g$. The goal of ICL is to learn a function that can utilize the context set $\gD$ to provide predictions for new queries $\vx_*$. To achieve this, the model is trained on different draws of context sets ($\gD_1, \gD_2, ...$) which share the same underlying functional mapping $g: \vx, \vz \to y$ but different realizations of the latent $\vz$, for example $g(\vx, \vz) = \vz^T\vx$ could be a linear regression system shared across different contexts $\gD_1, \gD_2, ...$, but the underlying latents could be different, i.e. $\gD_1$ is generated from $\vz_1$ while $\gD_2$ from $\vz_2$, similar to~\citet{pmlr-v202-von-oswald23a}. We emphasize that in this setup, we are not training models to do next-token prediction as is done in language modeling; instead, given a fixed context $\gD$ that includes $n$ samples, we are attempting to make a prediction on a single novel query $\vx_*$. We therefore do not use a causal Transformer, and we allow all tokens to attend to each other.

\looseness=-1
Often, ICL solutions are learned via maximum likelihood:
\begin{align}
\label{eq:mle}
    \arg\max_\varphi \;\;\mathbb{E}_{\gD, \vx_*, y_*}\left[\log p_\varphi(y_* | \vx_*, \gD)\right]
\end{align}where $p_\varphi$ represents the Transformer model and $\gD$ is sampled from the parametric family defined through $g$. Thus, the transformer model $p_\varphi$ must not only learn the form of the prediction function $g$, but also how to efficiently aggregate information from the context $\gD$ to infer $\vz$ for downstream predictions on $\vx_*$. Thus, this general framework can be naturally decomposed into two distinct parts.

\begin{figure*}
\centering
\includegraphics[width=\linewidth]{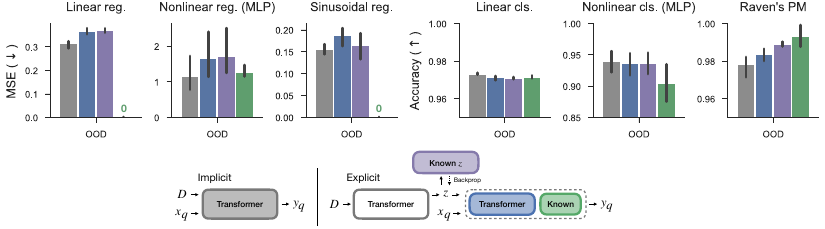}
    \vspace{-4mm}
    \caption{Performance on a subset of tasks where the true latents $\vz$ and prediction function $g$ are known. Implicit models are in shown \textcolor{gray}{gray}, explicit models with Transformer prediction in \textcolor{NavyBlue}{blue}, models trained with an auxiliary loss to predict the true latents in \textcolor{Purple}{purple} and those using the true prediction function in \textcolor{OliveGreen}{green}. Using the known prediction function leads to significantly better OOD performance.}
    \vspace{-6mm}
\label{fig:predictor}
\end{figure*}
\looseness=-1
\textbf{Context Aggregation.} This component deals with inferring the task-dependent latent variables from the in-context examples such that the downstream prediction becomes conditionally independent of the context, i.e. inferring $\vz$ from $\gD$ such that $p(y_* \mid \vx_*, \vz, \gD) = p(y_* \mid \vx_*, \vz)$.

\looseness=-1
\textbf{Predictive Modeling.} This component refers to the process of estimating the predictive function that leverages context $\gD$ to infer $y_*$ from a query $\vx_*$. In the above example, it refers to learning the functional mapping $g$ once $\vz$ has been extracted from context aggregation.

\looseness=-1
As discussed, Transformers do not have a clear incentive to make this explicit separation of context aggregation and predictive modeling. Instead, given context $\gD$, they implicitly and jointly model both the function $g$ along with $\gD$-dependent latent variable $\vz$ inference to directly provide predictions for the query $\vx_*$, in contrast to separately estimating $g$ and explicitly factorizing $\vz$. Thus, in order to enforce explicit representation of $\vz$, we propose a simple architectural modification where the query $\vx_*$ cannot directly attend to the context, and the latent task representation is forced to summarize the context efficiently. Formally, we compare the following two models, which are illustrated in \autoref{fig:explicit_implicit}.

\looseness=-1
\textbf{Implicit Model.} This refers to the traditional in-context learning computation performed by Transformer models. In this setup, given the set of observations $\gD$ (context) and a query $\vx_*$, the prediction $y_*$ is modeled directly as $p_\varphi(y_* | \vx_*, \gD)$, where $p_\varphi$ is defined using a standard Transformer with parameters $\varphi$ and is tasked with modeling both context aggregation and predictive modeling.

\looseness=-1
\textbf{Explicit Model.} This represents the architectural variation which minimally modifies the Transformer architecture by separating context aggregation and predictive modeling. It first constructs a task representation $\vz_\psi(\gD)$ using the set of observations $\gD$ and a context model $\vz_\psi$ with parameters $\psi$ (\textit{context aggregation}) and another network $p_\gamma$ to make a prediction for a new point $\vx_*$ (\textit{predictive modeling}) as $p_\gamma(y_* | \vx_*, \vz_\psi(\gD))$. A key insight is that the \textit{task latents} are invariant to the queries when modeling prediction. The context model is implemented with a Transformer $\vz_\psi$ with weights $\psi$, and for the prediction function $p_\gamma$, we study both Transformers and MLPs with weights $\gamma$. Importantly, the output of the context model $\vz_\psi(\gD)$ is a fixed-size vector with much lower dimensionality than the full context $\gD$. This information bottleneck prevents the query $\vx_*$ from attending directly to the context as in standard Transformers; instead, the context model must summarize $\gD$ into underlying generative factors, thus ruling out potential shortcut solutions that bypass latent variable inference.

\looseness=-1
\textbf{Implicit vs. Explicit.} Assuming Transformers do in fact favour shortcut-based solutions, we first hypothesize when each setup should perform better given different task characteristics. If the data is generated with a parametric model with few underlying parameters $\vz$ (e.g. a linear model $y=\vz^T\vx$), the right predictor can be precisely described using the parameters $\vz$, making the explicit model better suited. In contrast, if the data is generated with a Gaussian Process (GP), the implicit model should be superior since by construction query prediction computes similarities with all points in the context. In this case, the task latents of GP-based data with RBF kernel are infinite dimensional (i.e. a point in function space), which cannot be captured in the finite-dimensional bottleneck of the explicit model.
In general, we should expect the explicit model to be superior when the underlying true model is parametric and low-dimensional, but in case of a non-parametric or very high dimensional parametric model, the implicit model should be better. 
% Once again, these results are what we should expect to see only \textit{if Transformers bypass latent variable inference in favour of shortcut solutions} where the query attends directly to the context, which is our primary question of interest.

\looseness=-1
Finally, we note that our aim is \textit{not} to construct the best possible explicit model architecture -- indeed, more sophisticated ones already exist based on amortized Bayesian inference \citep{garnelo2018neural,mittal2023exchangeable}. Instead, we are interested in investigating potential inductive biases for ICL by \textit{minimally} modifying the standard Transformer architecture to remove certain shortcuts from the space of possible solutions. We leave the design of more performant architectures for future work and refer the readers to \autoref{apdx:related-work} for a detailed discussion of related work.

\begin{figure*}
\centering
\includegraphics[width=\linewidth]{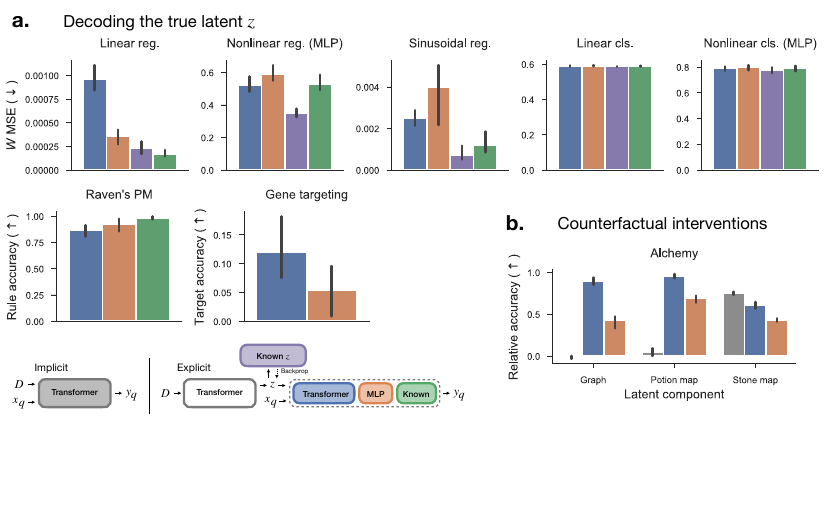}
    \vspace{-4mm}
    \caption{
    \looseness=-1
    Explicit models are interpretable as the bottleneck allows (a) linearly decoding the true latent, and (b) intervening to obtain correct counterfactual predictions. Implicit models are shown in \textcolor{gray}{gray}, explicit models with Transformer prediction in \textcolor{NavyBlue}{blue}, and with MLP prediction in \textcolor{Orange}{orange}. Models in \textcolor{OliveGreen}{green} use the true prediction function $g$, while models in \textcolor{Purple}{purple} use an additional auxiliary loss based on true latents. To evaluate decoding performance in (a), we linearly decode the true latent directly from concatenated context examples, which yields significantly worse performance than decoding from the bottleneck. Baseline performances in units of the plots are -- Linear regression: $0.49$, Nonlinear regression (MLP): $0.94$, Sinusoid regression: $0.33$, Linear classification: $0.86$, Nonlinear classification (MLP): $0.97$, Raven's PM: $0.5$, and Gene targeting: $0.0$. In (b), the ``Relative accuracy'' takes the baseline in account (details in \ref{section:das}).}
    \vspace{-6mm}
\label{fig:interpretability}
\end{figure*}
\vspace{-3mm}
\section{Experiments}
\vspace{-1mm}
\label{sec:experiments}
\looseness=-1
Our goal is to use both synthetic and real tasks that capture the key elements of ICL applications to tease apart the effects of implicit and explicit models on both in-distribution (ID) and out-of-distribution (OOD) generalization.

\looseness=-1
\textbf{Task Setup.} We conduct experiments across a variety of settings that admit task latents, from synthetic regression and classification to reasoning problems. For reasoning tasks that require compositional knowledge, we consider Raven's Progressive Matrices (Raven's PM)~\citep{John2003}, Alchemy~\citep{wang2021alchemy}, Gene Targeting~\citep{norman2019exploring} and reusable mixture of experts. A brief description of our tasks is provided below, with a more thorough explanation in \autoref{section:task}. 

\looseness=-1
\textit{Regression Tasks.} We consider different data-generating processes, e.g., linear: prediction $\vz^T\vx$ and latents $\vz$, nonlinear (MLP): prediction with a neural network $g(\vx, \vz)$ and latents as its weights, sinusoidal: prediction as a sum of sinusoids with different frequencies and the latents as its amplitudes.

\looseness=-1
\textit{Classification Tasks.} Akin to the regression problems, we consider a linear and nonlinear (MLP) prediction for classification using an additional sigmoid activation on the output.

\looseness=-1
\textit{Raven's Progressive Matrices.} A pattern-completion task used in IQ tests where individual object attributes evolve according to different rules. The task is to complete a sequence of frames to satisfy the underlying rule, which is the latent variable and can be based on colors, shapes, etc.
% such that the underlying rule, which is the latent variable and can be based on colors, shapes, etc., is maintained. 

\looseness=-1
\textit{Alchemy.} Here, a latent causal graph describes how different stones and potions interact to generate new stones. The task is to infer the next stone given some transitions.

\looseness=-1
\textit{Gene Targeting.} It represents a real-world dataset of targeted gene knockouts and subsequent observations of gene expressions across many cells. The underlying latent variable is the set of genes that were knocked out in a given experiment, and the task is to infer the gene expressions of certain cells in an experiment given those of other observed ones.

\textit{Reusable Modular Mixture of Experts (MoE).} In this task, we apply a sequence of operations $g_l$ on the input $\vx$, where the choice of expert applied at layer $l$ is governed by a categorical latent $\vz_l$. In particular, we apply five operations sequentially, leading to $y = g_{\vz_5} \circ g_{\vz_4} \circ \ldots \circ g_{\vz_1}(\vx)$. Here $g_1, g_2, \ldots$ are shared across tasks and each task is uniquely identified by $(\vz_1, \ldots, \vz_5)$ which are the task latents. This represents a reusable mixture of experts system with a hierarchical and compositional decomposition.

\looseness=-1
\textbf{Training and Evaluation.} Tasks based on regression utilize the mean-squared error loss, while those based on classification use the cross entropy loss for training. Model architecture details are provided in \autoref{sec:general_details}. For ID evaluation, we consider the underlying task latent $\vz$, context samples $\gD$, and queries $\vx_*$ to be sampled from the same distribution as during training. The challenge in this case is simply to generalize from finite data. For OOD evaluation, we test two different cases depending on the kind of task. For our synthetic regression and classification tasks, the task latent $\vz$ and context samples $\gD$ are sampled from the same distribution as at training time, but the queries $\vx_*$ are sampled from a Gaussian distribution with higher ($3\times$) standard deviation, thus testing a form of out-of-domain generalization. For our reasoning-based problems, we evaluate on task latents $\vz$ that were not seen at training. The task latent in each of these reasoning-based problems is composed of parts (e.g., in the Gene Targeting experiment, the latent is a binary vector of targeted genes), which allows us to test a form of \textit{compositional} generalization \citep{wiedemer2023compositional} in which all parts have full marginal support at training time, but novel combinations of those parts are evaluated at test-time.

\looseness=-1
For all tasks, implicit and explicit models were trained from scratch over a distribution of tasks latents, with a control for the number of parameters to provide a systematic comparison between the two. To better understand the shortcomings of different models, we also compare with privileged oracles (known decoder -- using ground-truth $g$ function, and known latent variable -- using an auxiliary loss that includes the ground-truth $\vz$). 

\looseness=-1
\textbf{Explicit models do not outperform the implicit models.} The first evaluation setting that we considered was the ID performance. In this case, we should expect both implicit and explicit models to perform equally well, even if implicit models learn shortcuts rather than ground-truth task latents. This is because those shortcuts are tuned to minimize prediction error within the same data distribution that is being evaluated. Across all our tasks, the results indeed confirmed this prediction. Specifically, during ID evaluation, all models were capable of making highly accurate predictions (\autoref{fig:results}). While the performance of the implicit model was generally slightly better than that of the explicit models, the benefits were marginal. Effectively, all models solved the tasks similarly well.

\begin{figure}
    \centering
    \includegraphics[width=\linewidth]{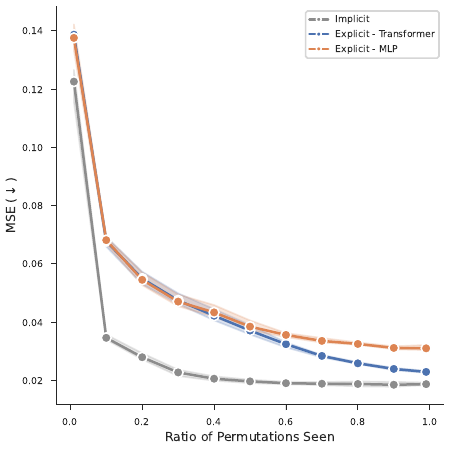}
    \vspace{-5mm}
    \caption{We conduct experiments on reusable modular MoE task where we train on a subset of combinations of experts, shown
    on the X-axis. Our results indicate that across different percentages of combinations seen during training, the implicit model consistently outperforms the explicit one at compositional generalization.}
    \label{fig:moe}
    \vspace{-7mm}
\end{figure}
\looseness=-1
While we expected that ID evaluation would be insufficient to demonstrate potential benefits of the explicit models, we expected to see differences in OOD settings. Both implicit and explicit models are sufficiently expressive to fit the training distribution. However, if an explicit model successfully learns the true task latents that generated the data while an implicit model learns statistical shortcuts that are specialized to the training distribution, we should expect the explicit model to generalize better OOD. As a reminder, for the synthetic regression and classification tasks in \autoref{fig:results} (a, b), OOD evaluation was done by sampling $\vx_*$ from a normal distribution with wider standard deviation than was used at training ($3\times)$, effectively evaluating if the models could extrapolate to points further out along their domain despite only being trained within a narrow distribution near the origin. For the compositional tasks in \autoref{fig:results} (c), we instead evaluated OOD performance by only training on certain configurations of the true latent variable $\vz$ while evaluating on unseen ones. Importantly, at training time the models were shown data that included every possible value for each component of $\vz$, but not every possible combination of these values were seen, thus evaluating a form of compositional generalization \citep{wiedemer2023compositional}. Similarly, for the reusable modular MoE task, we train the models on a fraction of all possible combinations in the latent $(\vz_1, \ldots, \vz_5)$ and evaluate on all combinations.

\begin{figure*}
\centering
\includegraphics[width=0.77\linewidth]{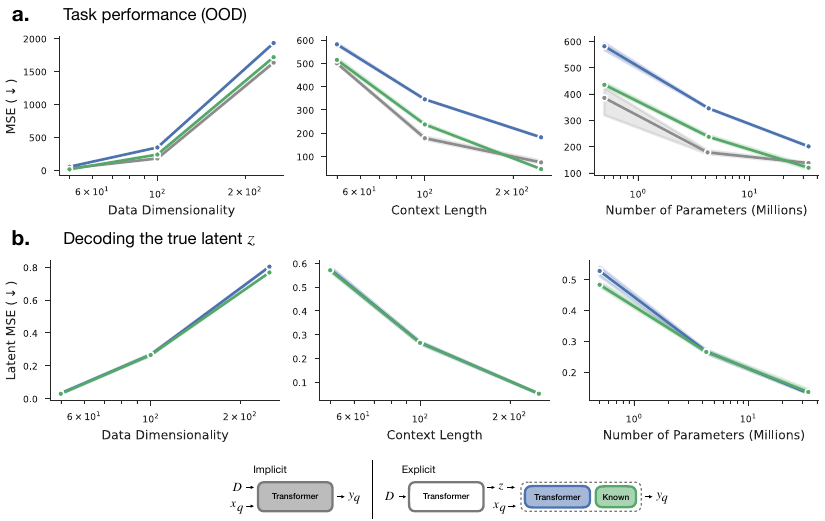}
    \vspace{-4mm}
    \caption{We analyze (a) Linear regression OOD performance and (b) latent variable linear decoding as a function of model and task parameters. Task performance scales similarly for implicit (\textcolor{gray}{gray}) and explicit models (\textcolor{NavyBlue}{blue}), while latent variable decoding scales similarly for the explicit model and models trained with the true prediction function $g$ (\textcolor{OliveGreen}{green}).}
    \vspace{-6mm}
    \label{fig:scale}
\end{figure*}
\looseness=-1
Surprisingly, and counter to our predictions above, we found that the explicit model provided no significant benefit in OOD settings. In synthetic regression tasks shown in \autoref{fig:results} (a), all models failed to generalize and obtained substantially worse performance than during ID evaluation, with the implicit model actually being the one that had a slightly lower degradation in performance. In classification and compositional tasks shown in \autoref{fig:results} (b, c), all models generalized fairly well OOD and with similar performance. Our results on reusable modular MoE task in \autoref{fig:moe} further indicate that implicit models consistently outperform explicit ones across different proportion of latent combinations seen during training (X-axis). In summary, explicit models appear to provide no benefit across our tasks, both in terms of ID and OOD performance.

\looseness=-1
If the explicit model did learn the right latent variables in the bottleneck, it essentially implies that either the implicit model learns benign shortcuts (if at all) or that learning the right latent variables is not sufficient to improve generalization, both ID or OOD. In the following results, we see that the explicit model does indeed learn the right task latents.

\looseness=-1
\textbf{Explicit models learn to infer the correct latent variable, but not how to use it.}
Why didn't the explicit model provide any benefit? Our initial hypothesis was that the implicit model could be susceptible to learning shortcuts that are sufficient to reduce the training loss and easy to express using self-attention between the query $\vx_*$ and context $\gD$. By summarizing the context in a bottleneck $\vz_\psi$, the explicit model should instead be forced to extract the true latent variable in order to minimize the training loss, thus learning a solution that is closer to the actual data-generating process. There are then two possible explanations for the results in \autoref{fig:results}: (1) the explicit models did not learn to extract the true latent variable despite inductive biases to do so induced by the bottleneck, or (2) they did extract the true latent variable but did not learn to use it in the correct way. We performed several experiments to test these two possibilities, and found strong evidence for the second.

\looseness=-1
To test whether or not the explicit models failed because they did not extract the correct latent variable, we added an additional supervised loss term to Equation~\ref{eq:mle}, $||\vz - W\vz_\psi||^2$, to force the explicit model to recover the true task latent $\vz$ up to a linear transformation, where $W$ is a learnable parameter.
% first attempted to encourage them to do so more directly by training an additional linear model which took $\vz_\psi$ as input and attempted to predict the true latents $\vz$. 
The loss of this linear model was then backpropagated through the context model along with the task loss. Results in \autoref{fig:predictor} (\textcolor{Purple}{purple}) show that this auxiliary loss provided no improvement apart from minor increases in accuracy on Raven's PM, suggesting that incorrect latent variable inference does not explain the explicit model's suboptimal performance. Indeed, when we did not use the auxiliary loss as a training signal for the explicit model and just evaluated the quality of the learned task latents by training a separate linear predictor to predict $\vz_\psi$, we found that we could still accurately predict the true latent variable (see \autoref{fig:interpretability} (a)). This means that in the absence of any explicit training signal to predict the true task latent, the explicit model accurately learns task latents up to linear reparameterizations.

\begin{figure*}
    \centering
    \includegraphics[width=\linewidth]{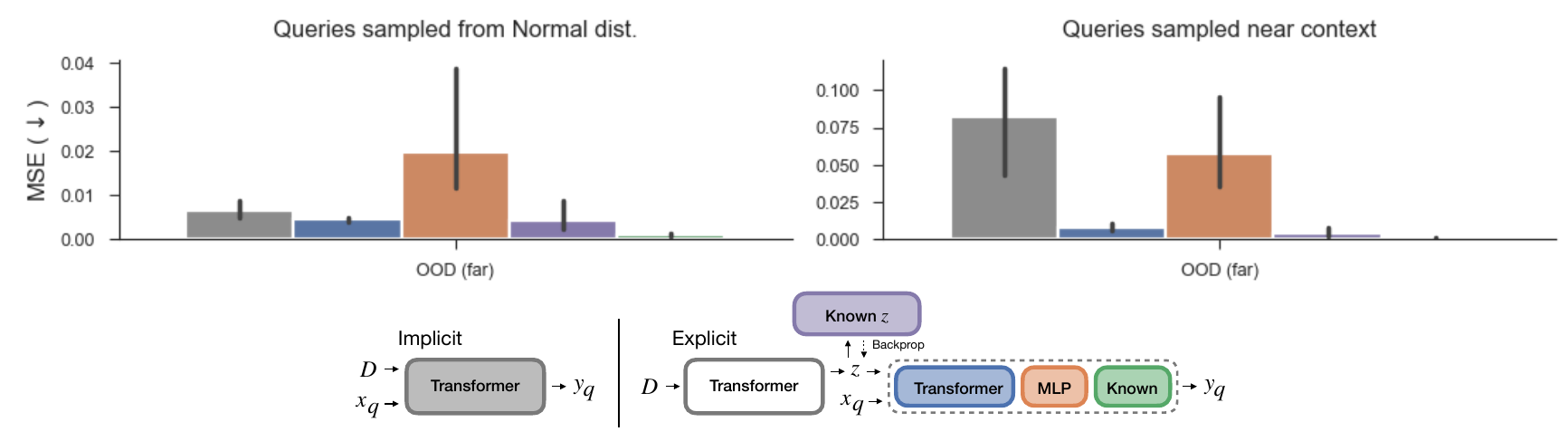}
    \vspace{-5mm}
    \caption{We analyze sinusoid regression by sampling the query either from the normal distribution (\textit{left}) or close to the context (\textit{right}) during training but always far from context during evaluation. We see that when queries are sampled close to the context, implicit models which can rely more on kernel-regression based nearest-neighbor solutions don't generalize far from context, while explicit models do.}
    \label{fig:shortcut}
    \vspace{-6mm}
\end{figure*}
\looseness=-1
Given that the explicit model correctly infers the true latent variable in its bottleneck, we study whether the prediction function is suboptimally learned. In other words, despite the explicit model having access to the true $\vz$, we hypothesize that $p_\gamma(y_* | \vx_*, \vz_\psi)$ does not learn the true data-generating process $y_* = g(\vx_*, \vz)$, where $g$ is the true prediction function -- e.g., for linear regression $g(\vx_*, \vz) = \vz^T \vx_*$. To validate this hypothesis, we trained explicit models with the prediction function $g$ hard-coded as an oracle. For instance, in the linear regression task, the $\vz_\psi$ output by the context model was linearly projected to the same dimensionality as the true weights $\vz$, after which the prediction function took its dot product with queried input $\vx_*$. In this setting, if the explicit model extracts the correct latent variable, it should generalize perfectly both in and out of distribution. Our results in \autoref{fig:predictor} confirm that using the correct prediction function indeed provides substantially better OOD generalization and effectively solves most tasks. This finding has significant implications: it suggests that while learning the true task latents may be a necessary condition for generalization, this must also be supplemented with significant inductive biases in the prediction function -- for instance, through novel architectures -- so that these task latents can be leveraged correctly. In the absence of such inductive biases, inferring the correct task latent appears to provide no benefits in practice. We do note that our nonlinear regression tasks, where $\vz$ represents the weights of an underlying MLP generating the data, were an exception to the results described here in that using an oracle prediction function performed poorly. In this case, we conjecture that the underlying latent variable is too difficult to accurately infer from the context, while shortcut-based solutions would avoid latent variable inference altogether to provide robust solutions.

\looseness=-1
\textbf{Explicit models are easily interpretable.}
While explicit models do not provide performance benefits, the ability to extract the correct latent and summarize it in a single bottleneck can still be useful for interpretability. On tasks with known underlying latent variables, we were able to linearly decode them from $\vz_\psi$ with high accuracy, meaning that the information is not only present but also easily accessible (\autoref{fig:interpretability} (a)). The exceptions were complex nonlinear regression tasks where the latents are difficult to infer and classification tasks where more context samples are needed to precisely identify the decision boundary. 
In contrast, finding such clear task-relevant latents is immensely challenging in an ordinary Transformer trained to do ICL, given that they can be distributed across a mixture of many layers and token positions.

\looseness=-1
Furthermore, even when latent variables appear to be successfully identified using a linear decoder in some hidden layer of a Transformer, one often finds that the relationships merely amount to spurious correlations \citep{ravichander2021probing}. For instance, if one manually modifies the activations in this hidden layer such that a different latent variable is predicted by the linear decoder, the model's predictions do not generally change in a way that is ``counterfactually'' consistent with this new latent (i.e., the prediction is not what it should have been under the new latent variable). We therefore used the Distributed Alignment Search (DAS) method from \citealt{geiger2023finding} (see \autoref{section:das}) to search for units in the implicit and explicit models that can be manipulated to obtain correct counterfactual predictions. For the explicit model, we limited this search to the bottleneck $\vz_\psi$. We found that using the explicit model, we were indeed able to manipulate $\vz_\psi$ and obtain correct counterfactual predictions, but we were not able to successfully do this using the implicit model, as shown in \autoref{fig:interpretability} (b). Explicit models might therefore be useful for both mechanistic interpretability and scientific discovery \citep{geiger2023causal}, where we do not know the underlying task latents and want to be able to easily uncover them from the trained model, and subsequently obtain a good predictor for an intervened system zero-shot given some knowledge about the intervention.

\looseness=-1
\textbf{Scaling Trends across Different Properties.}
To better compare the implicit and explicit models, we investigated their OOD task performance on linear regression as we varied the different properties of the task (input dimensionality and context length) and the size of the model used (\autoref{fig:scale} (a)). We found that performance in both models scaled similarly, but that the implicit model reliably outperformed the explicit one unless it used the known prediction function $g$. We also looked at the latent variable linear decoding accuracy in the explicit model as a function of these task and model properties (\autoref{fig:scale} (b)). As expected, we found that the latent variable was easier to decode from the explicit model's bottleneck when there was less inherit uncertainty about its value (lower data dimensionality, longer context length) and when the explicit model was given more capacity. However, throughout the different settings, we see that while the explicit model does learn the true latent well, it is not sufficient to get a performance boost over the implicit models. Further details on the setup of these scaling experiments is provided in \autoref{sec:scaling_details}.

\begin{figure}
  \begin{center}
    \includegraphics[trim={5mm 10mm 5mm 10mm},clip,width=0.8\columnwidth]{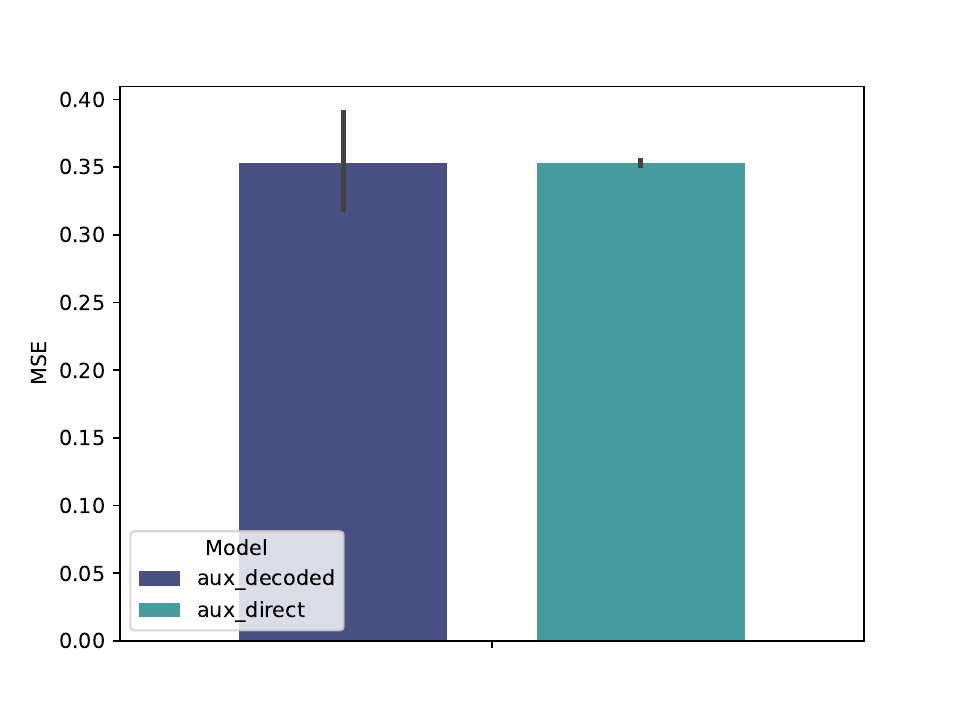}
  \end{center}
  \vspace{-5mm}
  \caption{We compare using the auxiliary ground-truth task latent loss directly on the output of context aggregation, i.e. $\vz_\psi(\gD)$ (aux\_direct), or to a linear decoding from it (aux\_decoding).}
  \vspace{-6mm}
\end{figure}
\looseness=-1
\textbf{Impact of auxiliary loss on decoding from bottleneck.} Additionally, we perform an experiment where instead of using an auxiliary loss obtained between the ground-truth task latents $\vz$ and a decoding from the explicit model's bottleneck $||\vz - W\vz_\psi||^2$ (called aux\_decoded), we instead force the bottleneck itself to be directly close to the ground-truth $||\vz - \vz_\psi||^2$ (called aux\_direct). Since the prediction function relies on the bottleneck and not its decoding, removing this extra layer when providing additional supervision might allow the bottleneck to better reflect the task latents and thereby aid prediction. Our results, however, indicate that doing so does not lead to any benefits on OOD evaluation for linear regression, further strengthening the conclusion that effective task latent inference is not the biggest problem in such models.

\textbf{Extreme Shortcut Injection.} Finally, we test whether injecting extreme shortcuts during training pushes implicit models to learn nearest-neighbor styled kernel-regression solutions as opposed to uncovering the underlying functional form. We consider two cases, where queries during training are sampled (a) randomly, or (b) near context points. The latter further incentivizes implicit models to learn nearest neighbour shortcuts. At evaluation, the queries are sampled far from the context. Our results on sinusoid regression in \autoref{fig:shortcut} indicate that while implicit models perform well generally, they suffer considerably more in the presence of such injected shortcuts since explicit models distill the task latent from context independent of the query.
% 
% To study this, we leverage the sinusoid task where we contrast training on queries sampled either randomly or close to some context point. At evaluation, the queries are sampled far from the context. Our results in \autoref{fig:shortcut} indicate that while implicit models perform well generally, they suffer considerably more in the presence of such injected shortcuts since explicit models distill the task latent from context independent of the query.

We further refer to \autoref{apdx:analysis} for a detailed analysis.
% additional analysis into our empirical results.

\vspace{-3mm}
\section{Conclusion}
\vspace{-2mm}
\looseness=-1
A commonly believed hypothesis is that Transformers do ICL through brittle statistical shortcuts rather than by inferring the underlying generative latent variables of the task, and that this explains their inability to generalize outside of the training distribution. Here, we empirically tested this hypothesis by minimally modifying the Transformer architecture through the use of a bottleneck that factorized the model into separate context aggregation and prediction functions, creating an inductive bias for explicit latent variable inference. While we confirmed that this model indeed learned to infer the correct latent variables across many ICL tasks, it surprisingly gave no improvement in performance for either in-distribution or out-of-distribution evaluation. Contrary to common belief, then, we showed that simply learning the correct latent variables for the tasks is not sufficient for better generalization because end-to-end optimization does not learn the right prediction model to leverage these latent variables.

% \clearpage
\section*{Impact Statement}
\looseness=-1
This paper provides a comparative analysis to better understand the capabilities of current in-context learners from a point of view of making them more robust and aligned with the true underlying models of the data. We believe that this is an important step towards understanding when and how machine learning models can rely on shortcuts and spurious correlations, and understanding whether such correlations can be mitigated through the use of conditional independence assumptions and bottlenecks, as is investigated in this work. 

While we show a negative result that such bottlenecks do not substantially aid generalization, they do come with interpretability benefits which are extremely useful when deploying AI systems at scale. Finally, we hope that our analysis sparks controlled experiments to understand the mechanisms behind in-context learning better as well as incorporating and validating numerous inductive biases to see whether they do aid generalization and reduce the reliance on shortcuts.

\section*{Ackonwledgements}
The authors would like to acknowledge the following researchers for valuable discussions and exchanges: Joao Sacramento, Johannes von Oswald. All authors acknowledge support from an unrestricted gift from Google inc.
EE acknowledges support from Vanier Canada Graduate Scholarship \#492702.
SM acknowledges the support of PhD Excellence Scholarship from UNIQUE.
DS acknowledges support from NSERC Discovery Grant RGPIN-2023-04869, and a Canada-CIFAR AI Chair.
GL acknowledges support from NSERC Discovery Grant RGPIN-2018-04821, the Canada Research Chair in Neural Computations and Interfacing, and a Canada-CIFAR AI Chair.

\clearpage
\bibliography{ref}
\bibliographystyle{icml2025}

\newpage
\appendix
\onecolumn

\section{Related Work}
\looseness=-1
\label{apdx:related-work}
\textbf{In-Context Learning.} In-Context learning (ICL) is an ability of certain trained models to take an entire task's dataset as input and parameterize solutions directly in their layer activations, which then condition subsequent computation on novel inputs from those same tasks. Generally, this ability is found in sequence models such as Transformers where the task dataset, or ``context'', corresponds to an earlier part of the sequence. ICL was first observed in large-scale pre-trained LLMs \citep{brown2020language}, and is similar in many respects to meta-learning \citep{chan2022data}. These LLM findings were subsequently expanded to more controlled settings outside of the language modality, where Transformer models were directly trained on task distributions such as linear regression \citep{pmlr-v202-von-oswald23a, garg2023Transformers}, hidden Markov models \citep{xie2022explanation}, compositional grammars \citep{hahn2023theory}, regular languages \citep{akyürek2024incontext} and Turing machines \citep{graumoya2024learning}, with a set of task observations defining the ``context''. These works highlight that Transformers are indeed able to model many types of complex task distributions, approaching in many cases the performance of the Bayes-optimal predictor, the a-priori optimal solution \citep{xie2022explanation}. Our work lies along similar lines but using more complex tasks and a systematic study into the differences between modeling the predictive space directly, or through a two-step process involving explicit inference of task latents.

\looseness=-1
\textbf{Shortcuts in ICL.} Shortcut learning is a phenomenon that has widely been observed in machine learning \citep{Geirhos_2020}, and refers to where a model solves a task through statistical correlations that are accidental and thus not robust to even slight distribution shifts. A classical example of this in image classification is the usage of background cues to classify objects \citep{ribeiro2016why}. Similar mistakes are know to be very common in NLP \citep{mccoy-etal-2019-right}, specifically in reasoning tasks \citep{zhang2022paradox}. Particularly relevant to our work, many authors have shown that has shown that Transformers are very prone to relying on shortcuts when performing ICL \cite{Tang_2023}. For instance, \cite{olsson2022context, singh2024needs} have shown that \textit{induction heads} play in important role in ICL by predicting that the continuation of a token will be the same as last time (i.e. $[a][b] \ldots [a] \rightarrow [b]$). As shown by \cite{pmlr-v202-von-oswald23a} this motif can be used to do linear regression, and can generally be seen as a form of kernel regression (i.e. $p(y_q|x_q,x_{1:n}) \propto \sum_i K(x_i,x_q)y_i$, \citep{han2023explaining}). This observation draws a link between those types of solutions and non-parametric inference methods in statistics \citep{nonparametric}, of which kernel regression is a member. In contrast \citep{hendel2023incontext, todd2023function} have concurrently shown that in some cases Transformers encode a ``task vector'' that they infer from the context and then use to do the prediction. There is therefore a need to better understand the nature of shortcuts in ICL and whether or not they can be easily avoided for better generalization. Our work explores this very question.

\looseness=-1
\textbf{Neural Processes.} 
The problem of solving new tasks in a zero-shot manner directly at inference is also closely tied to amortized Bayesian models \citep{kingma2013auto,rezende2014stochastic,radev2020bayesflow,geffner2023compositional,mittal2023exchangeable}. Conditional Neural Processes (CNPs)~\citep{garnelo2018conditional} provide a framework akin to the explicit model, where the posterior predictive distribution is modeled through a bottleneck $\vz_\psi$, i.e. $p_\mtheta(y_* | \vx_*, \gD) = p_\mtheta(y_* | \vx_*, \vz_\psi(\gD))$. However, CNPs do not look at the relevance of $\vz_\psi$ to the true latent $\vz$, and use the DeepSets~\citep{zaheer2017deep} architecture to model $\vz_\psi$, though recent research generalizes this setting to use Transformers \cite{nguyen2023Transformer} and other architectural backbones as well \citep{kim2019attentive,gordon2019convolutional}. Our approach with the explicit model, however, is to precisely question whether task-specific latents are encoded via $\vz_\psi$ which is now instead modeled using a Transformer architecture.  Analogously, Neural Processes (NPs) \citep{garnelo2018neural,pakman2020neural} augment CNPs with probabilistic modeling, where $\vz$ is now modeled explicitly as a latent-variable in the Bayesian sense, i.e. the likelihood is now modeled as $p_\mtheta(y | \vx_*, \gD) = \int p_\mtheta(y | \vx, \vz) p_\mtheta(\vz | \gD) d\vz$, where $\vz$ represents the latent variable and $\mtheta$ the parameters of the likelihood model. The model is trained via the Evidence Lower-Bound (ELBO) with the amortized variational approximation $q_\varphi(\cdot | \gD)$. Once trained, predictions for new datasets can be made by simply performing inference over the \textit{encoder} $q_\varphi$ to obtain $\vz$, and then leveraging this latent variable to eventually give the predictions via $p_\mtheta(y | \vx_*, \vz)$. Hence, while CNPs and the explicit model to share similarities in architecture, our goal is orthogonal in that we specifically use the explicit model to understand the impact of task-specific latent variable inference on ICL setups.

\looseness=-1
\textbf{Meta-Learning.} Meta-learning \citep{weng2018metalearning,hospedales2020metalearning} studies systems that can learn over two levels: rapidly through an inner-loop that is meta-learned using a slower outer-loop. The goal in such methods is to learn a good initialization common to the parameterized family of tasks, in a manner that obtaining a particular solution for a new task is fast from this initial point. The inner loop provides an optimization trajectory for a randomly drawn task from some initialization, which in itself is optimized in the outer loop to a good solution applicable for the global set of tasks. Typically, evaluation is done on some meta-validation set of tasks not seen during training. Task distributions can for example be a set of different classification/regression tasks (few shot learning, \cite{vettoruzzo2023advances}) or variations of a reinforcement learning (meta-RL, \cite{beck2023survey}). The goal is similar to ICL approaches in the sense that given a novel context $\gD$, one wants to make predictions for some query $\vx_*$. However, a big difference is that ICL approaches bypass modeling a common initialization by working directly on the prediction side (implicit), or instead predict the optimal parameters directly zero-shot through inference on the context model (explicit).

\looseness=-1
\textbf{Mechanistic Interpretability.}
Mechanistic interpretability is interested in understanding deep neural network's computations through interpretable abstraction, akin to what computational neuroscience does with the brain. \cite{alain2018understanding} introduced the foundational technique of linear ``probes'', which are linear models trained on the hidden state of a network to predict an abstract feature of the input; the success of which suggests that such a feature is used by the model. Since then, this naïve approach has been criticized for being potentially misleading \citep{ravichander2021probing}; in many cases a feature can be linearly decoded from a model without the model using it. More reliable methods grounded in causality \citep{vig2020causal, geiger2024finding} have now became the gold standard, and their use applied to Transformers has been exploding in popularity \citep{elhage2021mathematical}. 

\begin{figure}
    \centering
    \includegraphics[width=0.7\linewidth]{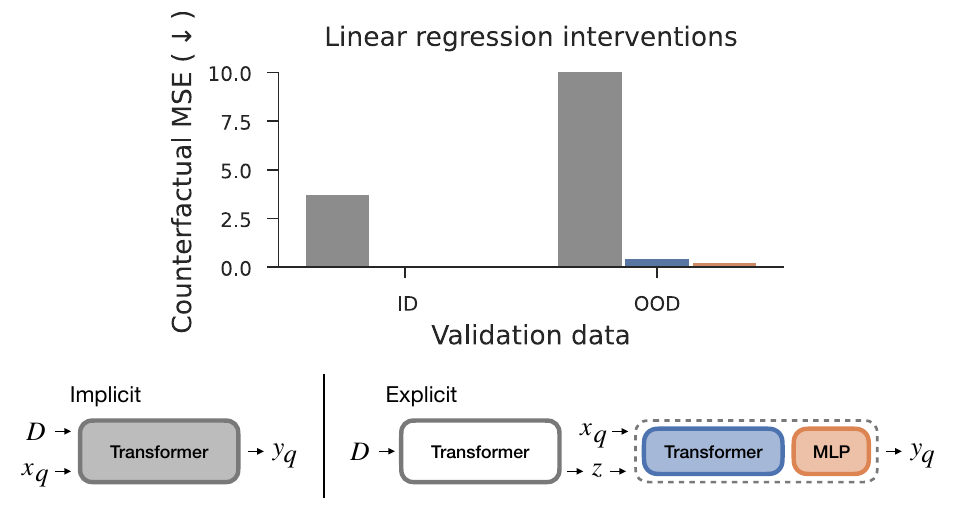}
    \vspace{-6mm}
    \caption{Same experiment as \cref{fig:interpretability}b but with the linear regression task. Specifically, we use Distributed Alignment Search (DAS, \cite{geiger2023finding}, see \cref{sec:model_details} for details) to find the 10 dimensional subspace in each model with best simulates counterfactual interventions on the task vector (in this case, the weight of the linear regression). In both explicit model, the subspace is taken at the bottleneck. In the implicit one, we perform the DAS at all layer of the query token and report the best one. The reported metric is the MSE of the intervened model on the intervened regression problem (i.e. using the same query $x$ but $y$'s coming from the intervened linear regression weights, \cref{sec:model_details} for details).}
    \label{fig:linear-intervention}
\end{figure}

\section{Tasks}
\label{section:task}
We consider the following tasks for our evaluations, specified by the data-generating prediction function $g: \vx, \vz \rightarrow y$ which is used to generate the ICL dataset, where $\vz$ represents the task-specific latent variable.

\subsection{Regression Tasks}\label{section:reg_task}
For regression tasks, since $y \in \mathbb{R}$, we use the mean-squared-error loss to train the model.

\textbf{Linear Regression.} This refers to the task where $y$ is obtained from an affine transformation on the input $\vx$. In particular, $\vy = g(\vx, \vz) = \vz^T \vx$, where $\vz\in \mathbb{R}^{i \times j} \sim \gN(0,I)$. For our experiments, we set $\dim(\vx) = 1$ and $\dim(\vy) = 1$.

\textbf{Nonlinear Regression using MLPs.} Here, the labels $y$ are obtained from a neural network which takes $\vx$ as an input. In particular, 
$\vy = g(\vx,\vz) = f_\vz(\vx)$, where $f_\vz$ is modeled as a Multi-Layer Perceptron (MLP) network with a $64$ dimensional single hidden layer and ReLU nonlinearity. The distribution of the weights of the neural network is $\vz \sim \gN(0, I)$. For our experiments, we set $\dim(\vx) = 2$ and $\dim(\vy) = 1$.
% The MLP has a shape of $[1,64,1]$ and ReLU non-linearities.
%Next, we look at systems where we define the prediction function as $y_i = f_\psi(\vx_i)$, where $f_\psi$ is modeled as a Multi-Layer Perceptron (MLP) network with weights $\psi$. This provides a more complicated problem since the space of task-specific parameters is much larger, equivalent to the parameter space in the defined MLP $f_\psi$. 
%Similar to the linear setup, we again consider regression with added noise. While knowing the parameters of $f_\psi$ is also maximally informative, these are higher dimensional and may not be perfectly represented by a typical reasonable bottleneck.

\textbf{Sinusoid Regression.} For this task, the label $y$ is obtained as a summation of sine functions with different frequencies and amplitudes, taking $\vx$ as an input. Mathematically, we structure the system as $y =  g(\vx, \vz) = \sum\limits_{i=1}^K \alpha_i \sin\left({2\pi \lambda_i \vx}\right)$, where $\lambda_i$'s denote the frequencies and $\alpha_i$'s the amplitudes. The parameters for the system can be seen as $\vz = \{\alpha_{1:K}\}$ while $\lambda_{1:K}$ remains fixed throughout. Additionally, for our experiments we set $K = 3$, and consider the distributions -- $\lambda_i\sim \mathcal{U}(0,5)$ and $\alpha_i\sim \mathcal{U}(-1,1)$, and set $\dim(\vx) = 2$ and $\dim(y) = 1$.

\textbf{Gaussian Process Regression.} While the other tasks considered had a parametric nature to it, this task on the other hand has more of a non-parametric nature. Here, the task is that $\vY \sim \mathcal{N}(\textbf{0}, K(\vX, \vX))$, i.e. the set of labels is sampled from a joint Gaussian distribution, akin to drawing a random function through a Gaussian Process (GP) prior and then evaluating it at different points $\vX$; with $K$ defining the kernel in the GP. In our case, we consider $K(\vx,\vx')=\exp\left(-\frac{\Vert \vx - \vx' \Vert^2}{2\sigma^2} \right)$ as the RBF kernel and $\vX  = (\vX_c, \vX_q), \vY = (\vY_c, \vY_q)$ are the combined points for both the context and the queries, which are split after this sampling. Here the latents $\vz$ has to store the kernel computations between the query and all the context points $\vX_c$, as well as the corresponding context labels $\vY_c$. Storing this either involves storing the high-dimensional mapping of $\vX_c$ which is defined by the kernel $K$, or storing all the points $\vX_c$ themselves. This is thus very high dimensional and weakly structured.

%\sar{add equations for GP} Based on our hypothesis, GPs provide a setup where the implicit model should be superior since by construction query prediction relies on computing its similarities with every points in the context, and thus a bottleneck would not be able to compress all of the context information. Additionally, it requires computing the notion of similarity through $k(\vx_*, \vx_i)$, which in some cases (RBF Kernel) requires computing the dot-product in an infinite dimensional space, since it only becomes easily tractable when \textit{provided} with the query. Thus, the sufficient statistics for the task is incredibly high-dimensional, and thus we hypothesize implicit models outperforming the explicit counterparts.

\textbf{Hodgkin-Hoxley ODE Prediction.} This is an example of the task where the context $\gD$ is not composed of \textit{iid} entries, but instead observations from the Hodgkin-Huxley temporal dynamics model of neural activity unrolled through time : 
$$C_m\frac{dV}{dt}=g_1\left(E_1-V\right) + {\bar{g}_{Na}}m^3h\left(E_{Na}-V\right) + {\bar{g}_{K}}n^4\left(E_K-V\right) + \bar{g}_Mp\left(E_K-V\right) + I_{inj}+ \sigma\eta\left(t\right)$$
Above, $V$ represents the membrane potential which is the target of interest, $\vt$ represents the different points at which observations are provided, $C_m$ is the membrane capacitance, $g_{\text{l}}$ is the leak conductance, $E_{\text{l}}$ is the membrane reversal potential, $\bar{g}_c$ is the density of channels of type $c$ ($\text{Na}^+$, $\text{K}^+$, M), $E_c$ is the reversal potential of $c$, ($m$, $h$, $n$, $p$) are the respective channel gating kinetic variables, and $\sigma \eta(t)$ is the intrinsic neural noise. The right hand side of the voltage dynamics is composed of a leak current, a voltage-dependent $\text{Na}^+$ current, a delayed-rectifier $\text{K}^+$ current, a slow voltage-dependent $\text{K}^+$ current responsible for spike-frequency adaptation, and an injected current $I_{\text{inj}}$. Channel gating variables $q$ have dynamics fully characterized by the neuron membrane potential $V$, given the respective steady-state $q_{\infty}(V)$ and time constant $\tau_{q}(V)$ (details in \cite{pospischil2008minimal}).

Importantly, in our experiments, we fix all parameters but $(\bar{g}_{Na}, \bar{g}_K)$ to values in \citealt{tejerocantero2020sbi} and solve the differential equation for 6,400 pairs $(\bar{g}_{Na}, \bar{g}_K) \in [0,40]^2$ from $t=0$ to $t=120$ with $1000$ time-steps. In other words, the Transformer has to regress to solutions of ordinary differential equations, where the task latents are $\vz = \{\bar{g}_{Na}, \bar{g}_K]\}$, the observations are $\vx = \vt$ and $y = V$, such that $y = g(\vx, \vz)$. Here $g$ represents the unrolling of the differential equation.

%It serves as another type of latent variable model where the latent is low-dimensional; we take two of the parameters of the ODE vary them as $\vz$.

\subsection{Classification Tasks}\label{section:cls_task}
For classification tasks, since $y$ is a categorical measure, we use a cross-entropy loss for training.
%Analogous to regression tasks, here the models see a set of observations $\gD = \{(\vx_i, y_i)\}_{i}$ and are tasked with classifying a new point $\vx_*$ with a cross-entropy loss.

\textbf{Linear Classification.} Akin to linear regression, here we consider the case that $y$ is obtained by an affine transformation of $\vx$ followed by a sigmoid function and a consequent sampling step. That is, $y = g(\vx, \vz) \sim \mathrm{Categorical}(\text{Softmax}(\vz^T \vx))$ where $\vz \in \mathbb{R}^{i \times j}\sim \mathcal{N}(0,I)$. For our experiments, we set $\dim(\vx) = 2$ and $y \in \{0, 1\}$.

\textbf{Nonlinear Classification Using MLPs.} Here, the logits for the labels are instead obtained through a neural network taking $\vx$ as an input, and not an affine transformation. Mathematically, this can be seen as
$y = g(\vx, \vz) \sim \mathrm{Categorical}(\text{Softmax}(f_\vz(\vx)))$ where $f_\vz$ is modeled as a Multi-Layer Perceptron (MLP) network with a $64$ dimensional single hidden layer and ReLU nonlinearity. The distribution of the weights of the neural network is $\vz \sim \gN(0, I)$. For our experiments, we set $\dim(\vx) = 2$ and $y \in \{0, 1\}$.

\subsection{Compositional tasks}\label{section:compositional}

\textbf{Reusable Modular Mixture of Experts (MoE).} We consider a modular task which consists of sequential application of a choice of $K$ experts $g_1, \ldots g_K$ over the input $\vx$. In particular, the computational graph consists of $L$ layers where at each layer $l$, expert $\vz_l$ operates on the output of the preceding layer to give the successive output, i.e. $\vx^l = g_{\vz_l}(\vx^{l-1})$. This task is compositional in nature because at each layer, any of the $K$ experts can be called upon to perform a unit of computation and the choice of the expert is defined by the underlying task latent $\vz_1, \ldots, \vz_L$, each of which are categorical with $K$ possibilities. In our specific implementation, we set $L$ to be 5, $K$ to be 5, $\vx \in \mathbb{R}^4$ and $\vy \in \mathbb{R}^4$. Each expert $g_i$ is parameterized as a linear layer followed by the tanh activation function. We enumerate all $K^L$ possible combinations and then only use a subset of them during training, while randomly sampling all for evaluation.

\textbf{Alchemy.} Alchemy is a meta-reinforcement learning benchmark \cite{wang2021alchemy} where each environment is defined by a set $\mathbf{z}=(\textsc{Graph, Potion map, Stone map})$ of rules about how some set of potions transforms some stones. We extracted from it an ICL classification dataset consisting of transformations $\mathbf{x}=(\textsc{stone}, \textsc{potion}) \rightarrow \mathbf{y}=\textsc{stone}$. The transformations are compositional and symbolic; each potion affects only one of the three properties of stones (size, shape and color). An environment is specified by how observable stones and potions \textsc{map} to latent stones and potions, along with a \textsc{graph} over these latent stones which specify the result of the Transformations. In total there is 109 \textsc{Graph}, 48 \textsc{Potion map} and 32 \textsc{Stone maps}, making for 167424 environments. We reserve 100,000 environments for evaluation and train of the remaining ones.

\textbf{Raven's Progressive Matrices.} Raven's Progressive Matrices (Raven's PM) is a reasoning task used for IQ tests \citep{John2003}. It consists of a 3x3 grid where each cell contains simple objects varying in a small number of attributes (number, shape, size, color), but the bottom right cell is left empty. Subjects must notice a pattern in how the cells change from left to right in the first two rows of the grid, and then use that same pattern to complete missing cell in the bottom row. This is done by selecting one answer among $N$ possible provided options for the missing cell. We use a symbolic version of the dataset that addresses bias in the original version \citep{guo2023emergent}. In this dataset, objects at a cell have $4$ discrete attributes with $40$ possible values each. In our models, the context consists of the first two rows of the grid, the query consists of the last row with a masked out final cell, and the ground-truth latent variable is the underlying rule that generates a particular grid. Each rule is composed of a set of sub-parts, and we evaluate on unseen compositions.

\textbf{Gene Targeting.} We use Perturb-seq dataset collected by \citet{norman2019exploring} where researchers performed several genetic intervention experiments using CRISPR \citep{gilbert2014genome}. In each experiment, either one or two genes were targeted and the resulting expressions across 5000 genes were observed across several cells. Here, we consider each CRISPR intervention experiment as a different context, the resulting cell genetic expressions as 5000-dimensional observations, and a left-out cell with half of the genetic expressions randomly masked out as the query. The task is to predict the missing genetic expressions for the queried cell. We evaluate on held on held out CRISPR experiments with novel pairs of targeted genes.

\section{Model Details}
\label{sec:model_details}
In the following section, we describe the standard architectural details used for all the tasks, as well as specific differences in the architecture used for the scaling experiments. Finally, we also provide details about the distributed alignment search mechanism.

\subsection{General Details}
\label{sec:general_details}
For our implicit model, we use a standard Transformer with 8 layers. In the explicit model, for context aggregation we parameterize $\vz_\psi(\gD)$ using a standard Transformer with 4 layers, 256 dimensions latent, 512 dimensions MLP, and 4 heads. For the predictor $p_\gamma$, we consider two options: a ReLU-actiavtion based MLP with three hidden layers of size 512 and a Transformer with the same configuration as $\vz_\psi(\gD)$. 

% How we feed data to models
For the implicit model, we format the prompt for prediction as $[\vx_1,y_1] \ldots [\vx_n,y_n] [\vx_q, \emptyset]$, where every $[\cdot]$ represents a token. We use a distinct mask token $\emptyset$ to represent the target (which is the thing being predicted). For the explicit model, we first compute $[\vx_1,y_1] \ldots [\vx_n,y_n]$ to $\vz_\psi(\gD)$ with the context Transformer, then we give $[\vz_\psi(\gD)][\vx_q]$ to the predictor Transformer or $[\vz_\psi(\gD), \vx_q]$ to the MLP. 

For our experiments, the number of context points $n$ is uniformly sampled from $16$ to $128$ for both training and evaluation. Training is done with new data being synthetically generated on the fly, and evaluation either based on the test set provided for real-world tasks or simulated data of $1000$ different contexts for synthetic tasks. All the models were trained with a learning rate of $10^{-4}$ using the Adam optimizer~\citep{kingma2014adam} for a $1000$ epochs.

\subsection{Scaling Experiments}
\label{sec:scaling_details}
For the scaling experiments, we only consider the linear regression case with a base configuration of: (a) $\vx$ of dimensionality $100$, (b) context size being sampled uniformly from $(75, 125)$, and (c) $8$ heads, $8$ layers, $512$ hidden dimensions and $256$ bottleneck dimension for the transformer models.

From this base configuration, we changed only one of the configurations at each time to test for scaling trends for each property independently. In particular, we ablated over $(50, 100, 250)$ for the dimensionality of $\vx$, $(50, 100, 250)$ for the context length which was sampled from a $\pm 25$ range and the model size. The smallest model size considered had $4$ heads, $4$ layers, $256$ hidden dimensions and $128$ feature dimensions. The medium multiplied each of these properties by $2\times$, and the biggest model subsequently multiplied it by $2\times$ again. For the explicit models, we considered the same scaling paradigms with the number of layers being split by half to accommodate a separate context model and prediction model.

All the models were trained with a learning rate of $10^{-5}$ using the Adam optimizer for $5000$ epochs.

\subsection{Compute Details}

We train most of our models on single RTX8000 NVIDIA GPUs, where it takes roughly 3-6 hours for each experiment to run. Our scaling experiments on the other hand often required 1-2 days on single GPUs for training each model.
\begin{figure}
    \centering  
    \includegraphics[width=\linewidth]{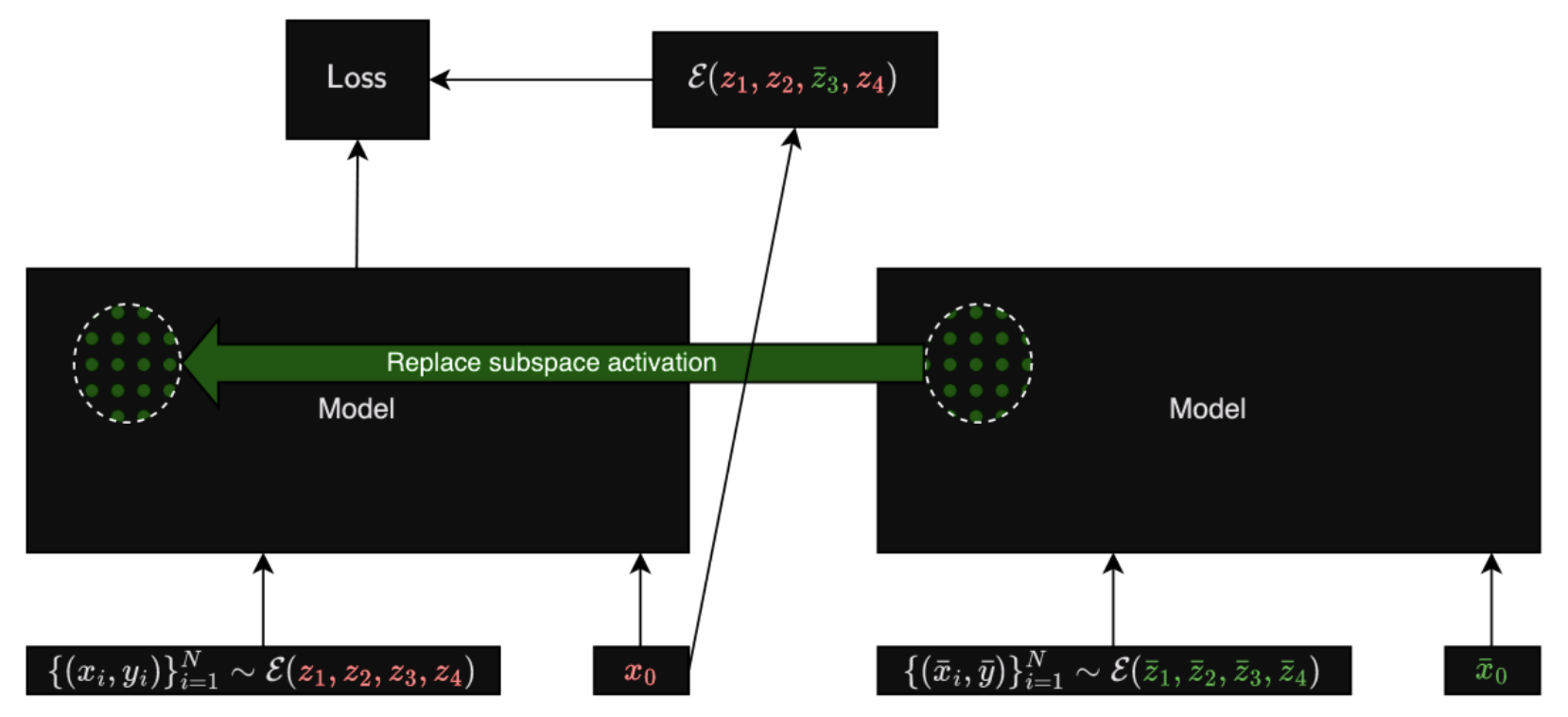}
    \caption{Illustration of the DAS training procedure}
    \label{fig:das}
\end{figure}

\subsection{Distributed Alignment Search details} \label{section:das}
To find subspace causally associated with a task latent in Alchemy, we use a method based on Distributed Alignment Search (DAS) by \cite{geiger2023finding}. This procedure is performed for a location $L = \mathbb{R}^d$ (e.g. the bottleneck) and latent $i\in \{1,2,3\}$ (\textsc{Graph, Stone map, Potion map}).

First, we run with the model on $\mathcal{D}_z$ and $\mathcal{D}_{\bar{\vz}}$ for every possible query $x_*$. We call $\vz$ the base and $\bar{\vz}$ the source and only differ by the $i$th latent. For every run, we record the activity of the source model at the location $l_z\in \bar{L}$. Then, we run the base model again but this time fixing the subspace of $l$ defined by the orthogonal projection $\Pi \in \mathbb{R}^{d\times 10}$ to it's value in $l_z$. A single projection $\Pi$ is learned over all possible combination $\vz, \bar{\vz}$ and $x_*$ with a cross-entropy loss between the prediction of the base (intervened) model and the true counter-factual result of changing the latent $z_i$ to $\bar{\vz}_i$. See \autoref{fig:das} for an illustration of the process. A subspace is evaluated by looking at the accuracy of the counterfactual interventions over a dataset of held-out $\vz,\bar{\vz}$ pairs; a quantity called the Interchange Intervention Accuracy (IIA). In Figure \autoref{fig:interpretability} (b) we report the validation IIA relative to a baseline corresponding to the counterfactual accuracy if we don't perform any intervention (because changing the latent sometimes doesn't change the prediction) $\frac{\textsc{IIA}-\textsc{baseline}}{1-\textsc{baseline}}$. 

\begin{figure}
    \centering
    \includegraphics[width=\linewidth]{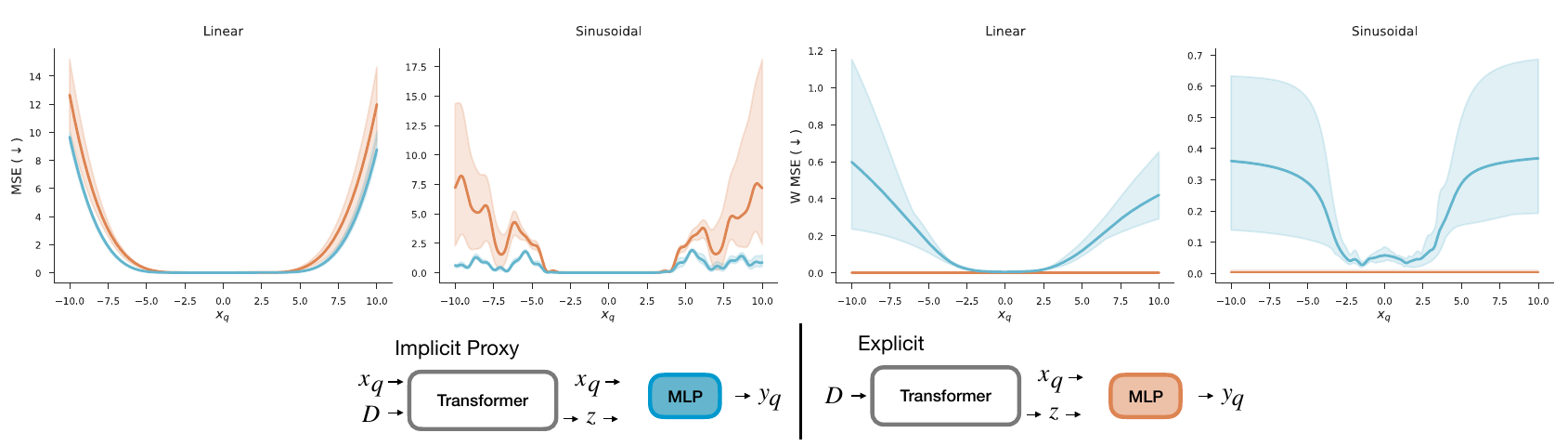}
    \vspace{-6mm}
    \caption{To further understand the difference between the explicit and implicit model, we utilize an implicit proxy model which shares the same architecture as the explicit model with MLP prediction with just \textit{one key difference}: the task latent $\vz$ depends on the query $\vx_q$ as well. This task latent $\vz$ can be understood as the final attention layer output of the implicit model, after which an additional MLP is utilized to provide prediction. Our findings on linear and sinusoidal regression demonstrate that as we move further and further out-of-distribution, the implicit proxy model performs better than the explicit model (left figures), but recovers the underlying true task latents worse (two right figures). This provides additional validation of our hypothesis.}
    \label{fig:ablation}
\end{figure}

\section{Analysis of Experiments}
\label{apdx:analysis}
Based on the empirical evidence presented in \autoref{sec:experiments}, we finally provide details and analysis into the results to further the understanding of the conclusions. In particular, our key analysis includes

\textbf{Explicit Models sufficiently uncover task latents.} We see that in problems where the context provides enough evidence to uncover the true task latents, explicit models are able to do so. In particular, this hints at the fact that explicit models do perform downstream prediction based on true task latents whenever these latents can be sufficiently identified from the context examples.

\textbf{Explicit Models do not generalize better than implicit ones.} Our analysis also reveals that while explicit models often do uncover the right task latents, they are still not able to surpass implicit models even on OOD generalization. This could be due to implicit models also uncovering the true underlying prediction function but in a distributed fashion, or explicit models not being able to leverage the learned latents in downstream prediction. 

\textbf{Learned downstream prediction is often sub-optimal.} Our results indicate that it is indeed the case that while the explicit models do uncover the right latents, they fail to generalize well OOD because the downstream prediction function fails to generalize.

This is further strengthened by \autoref{fig:ablation} where additionally leverage the query in context aggregation, thus interpolating between explicit and implicit model while maintaining a bottleneck. Our results indicate that despite worse latent variable inference far from the query, it still has better predictive generalization when compared to explicit models. This further strengthens the claim that better latent variable inference is not the sole problem, and learning the right downstream prediction is as important.

\textbf{Classification tasks vs. regression tasks.} OOD performance is generally strong (across all models) for classification because decision boundaries are within the training domain and do not change beyond it. In contrast, for regression tasks, the function continues to change beyond the observed training domain, making OOD prediction more difficult. This is also why known prediction functions give little benefit in classification tasks: they are already solved well OOD with ordinary implicit and explicit models.

\textbf{The explicit model with known prediction function does not give benefits in nonlinear (MLP) regression.} This is because the problem of inferring an MLP’s weights given some context examples is too difficult, so the explicit model opts for a different, non-parametric solution. This is supported by the latent variable decoding results in Fig. 5 (previously Fig. 4), which show that even with a known prediction function the explicit model does not learn to infer the correct latent variable for the nonlinear (MLP) regression task.

\textbf{Impact of Output Dimensionality}. In addition to our standard experiments, we consider linear regression with varying output dimensionality as another measure of difficulty of the task. Our results in \cref{fig:out_scale} showcase that the implicit model fares much better than the explicit model, however having a known prediction function leads to much better performance.

\begin{figure*}
    \centering
    % \vspace{-1mm}
    \includegraphics[width=\textwidth]{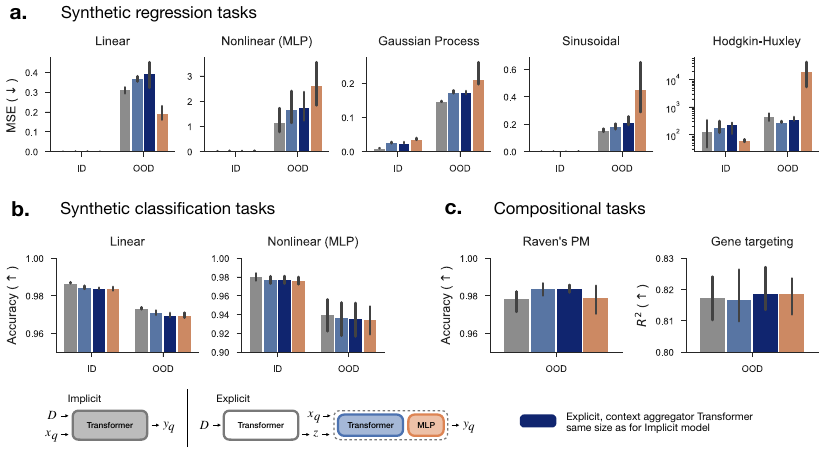}
    % \vspace{-9mm}
    \caption{Similar to Figure 2, but with an additional explicit model where the context aggregator is identical in size and hyperparameters (M) as the implicit model. Implicit models are in shown \textcolor{gray}{gray}, explicit with Transformer prediction in \textcolor{NavyBlue}{blue} (light blue = total model size identical to implicit as in main paper, dark blue = context aggregator identical to implicit), and with MLP prediction in \textcolor{Orange}{orange}.}
    \label{fig:apdx_size_1}
\end{figure*}

\begin{figure*}
    \centering
    % \vspace{-3mm}
    \includegraphics[width=\textwidth]{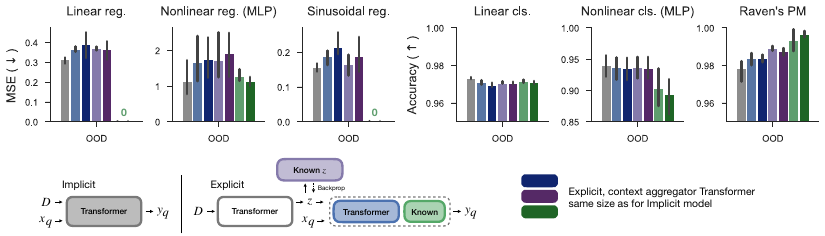}
    % \vspace{-8mm}
    \caption{Similar to Figure 3 but with an additional explicit model where the context aggregator is identical in size and hyperparameters (M) as the implicit model. Performance on tasks where the true latents $\vz$ and prediction function $g$ are known. Implicit models are in \textcolor{gray}{gray}, explicit models with Transformer prediction in \textcolor{NavyBlue}{blue}, models trained with an auxiliary loss to predict true latents in \textcolor{Purple}{purple} and those using the true prediction function in \textcolor{OliveGreen}{green}. Using the known prediction function leads to significantly better OOD performance. Lighter color indicates total model size identical to implicit as in main paper, darker = context aggregator identical to implicit.}
    \label{fig:apdx_size_2}
\end{figure*}

\begin{figure*}
    \centering
    \includegraphics[trim={13cm 0 13cm 0},clip,width=0.6\textwidth]{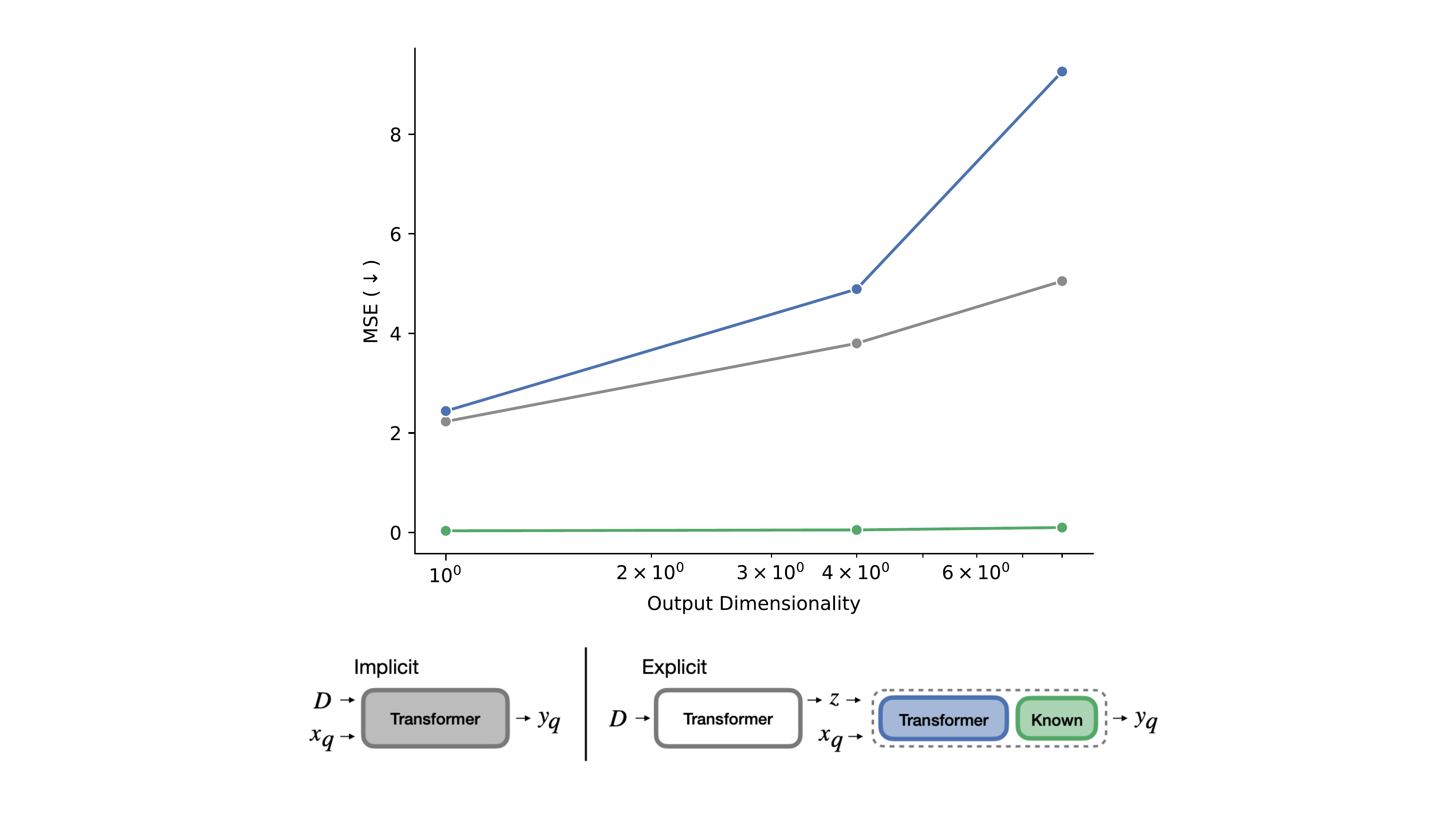}
    \vspace{-9mm}
    \caption{We study the impact of output dimensions on the performance of explicit and implicit models for linear regression with $8$-dimensional inputs. The output dimensionality is considered to be $1$, $4$ and $8$ dimensional, and our results indicate that implicit methods (\textcolor{gray}{gray}) outperform explicit ones (\textcolor{NavyBlue}{blue}), but the same explicit models with known prediction function (\textcolor{OliveGreen}{green}) scales much better.}
    \label{fig:out_scale}
\end{figure*}

\textbf{Size of Context Aggregator}. While we primarily controlled for the total number of parameters, it is possible that context aggregation requires more parameters and thus equally splitting parameters between this aggregation and prediction may be the cause for suboptimal performance. To study this properly, we conduct experiments where the context aggregator is the same size as the implicit model and we see similar results in \cref{fig:apdx_size_1,fig:apdx_size_2}. In addition, we ablate with differently sized context aggregators and prediction models for different tasks in \cref{tab:apdx_ablate_1,tab:apdx_ablate_2,tab:apdx_ablate_3,tab:apdx_ablate_4,tab:apdx_ablate_5,tab:apdx_ablate_6,tab:apdx_ablate_7,tab:apdx_ablate_8}

\begin{figure*}
    \includegraphics[width=\textwidth]{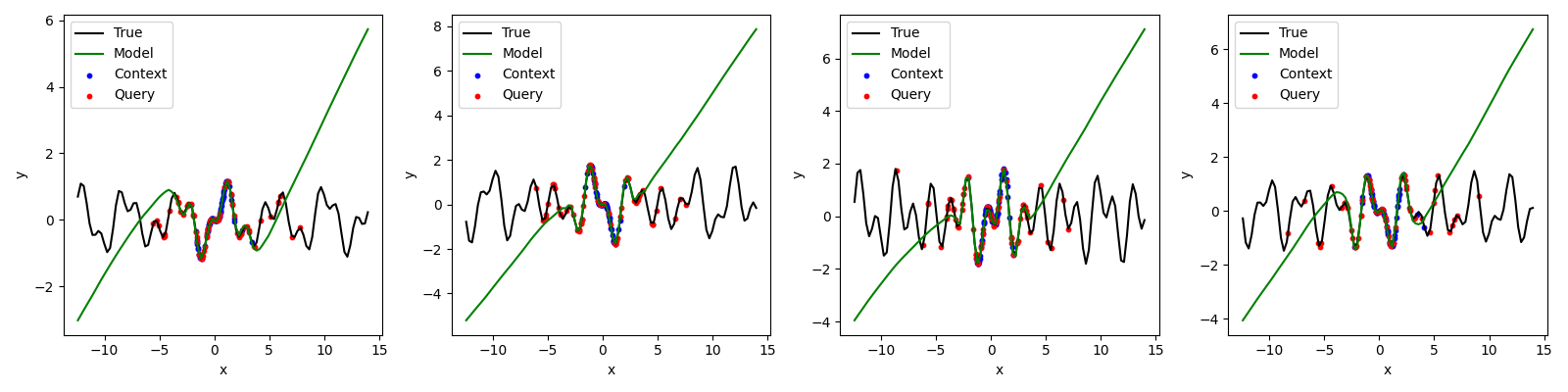}
    \vspace{-9mm}
    \caption{Illustration of explicit model with MLP prediction on the sinusoid task OOD. True function is shown in black, model output in \textcolor{OliveGreen}{green}, context points in \textcolor{NavyBlue}{blue} and query points in \textcolor{red}{red}. Our results indicate the failure of learned prediction function away from context.}
    \label{fig:sin_exp_vis}
\end{figure*}

\textbf{Visualization of Explicit Predictions}. We visualize the failure of explicit model in learning the right prediction function, which can be seen in out of distribution for sinusoid regression in \cref{fig:sin_exp_vis}.

\section{Mathematical Formalism}
\label{section:math}
In this section, we provide a formal distinction between the implicit and explicit model. In both the approaches, the goal is to model the true posterior predictive $p(y | \vx, \gD)$; however the two methods model it through different conditional independence setup. 

\textbf{Implicit Model}. In this setup, we model the predictive distribution as $p_\varphi(y | \vx, \gD)$, where the training is done as
\begin{align}
    \arg\max_\varphi \mathbb{E}_{\vx, y, \gD}\left[\log p_\varphi(y | \vx, \gD)\right]
\end{align}
and then given a query $\vx$ and dataset $\gD$, the inference is done simply by sampling or estimating the mean of $p(y | \vx, \gD)$.

\textbf{Explicit Model}. Contrary to the implicit model, the explicit model parameterizes the predictive distribution as $p_\gamma(y | \vx, \vz_\psi(\gD))$, with a similar training procedure as above. Note that the predictive distribution only interacts with the dataset $\gD$ through the latent $\vz_\psi(\gD)$ while the implicit model allows unconstrained access to $\gD$.

\textbf{Implicit Proxy Model}. To better understand the differences that play a role from architectural differences and parameterizations, we use exactly the same architecture as the explicit model to obtain a version of the implicit model. Such a model parameterizes the predictive distribution as $p_\gamma(y | \vx, \vz_\psi(\gD, \vx))$, with a similar training procedure as above. Note that the only difference with the explicit model here is that the conditional dependence of the query and the task latents is broken.

\begin{figure}
    \centering
    \includegraphics[width=\linewidth]{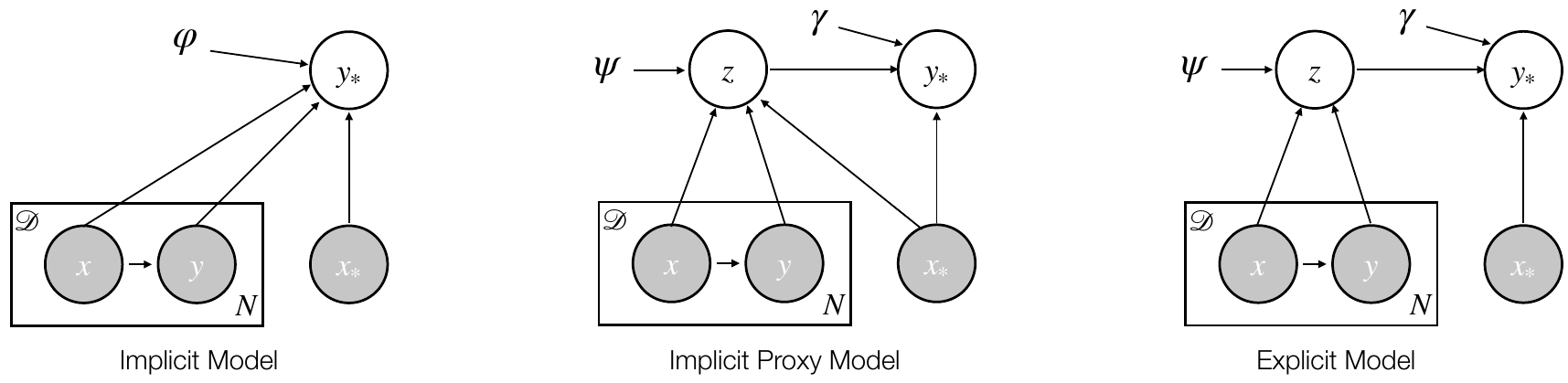}
    \caption{Plate diagram for the implicit model (left), implicit proxy model (middle) and the explicit model (right), where gray blocks refer to observed variables and white refers to unobserved variables. Trainable parameters are indicated without circles. In the explicit model case, $\vz$ is currently modeled as a dirac measure defined via the trainable parameters $\psi$. One can see the implicit proxy model as very similar to the implicit model where output of the last attention layer corresponding to the query token is further processed to give prediction. Its similarity to the explicit model is also clear as it shares exactly the same parameterization.}
    \label{fig:plate}
\end{figure}
We refer the reader to \autoref{fig:plate} for a plate diagram of the corresponding architectures.

\begin{table*}[]
    \centering
    \footnotesize
    \begin{tabular}{ccccccc}
    \toprule
    & & & & \multicolumn{3}{c}{\textit{$L_2$ Loss $(\downarrow)$}} \\
    \cmidrule{5-7}
    \textbf{Model} & \textbf{\# Context Layers} & \textbf{\# Prediction Layers} & \textbf{\# Parameters } (M) & \textbf{IID} & \textbf{OOD Query} & \textbf{OOD Latent} \\
    \midrule
\multirow{3}{*}{Explicit-Known} & 4 & - & 2.1 & \textbf{0.011} & \textbf{0.016} & 0.782 \\
& 6 & - & 3.2 & 0.012 & 0.019 & 0.923 \\
& 8 & - & 4.2 & 0.013 & 0.023 & 1.391 \\
\cmidrule{2-7}
\multirow{9}{*}{Explicit-MLP} & 4 & 4 & 3.0 & 0.015 & 0.169 & 0.883 \\
& 4 & 6 & 3.6 & 0.017 & 0.174 & 0.728 \\
 & 4 & 8 & 4.1 & 0.015 & 0.082 & 0.929 \\
 & 6 & 4 & 4.1 & \textbf{0.011} & 0.201 & 1.047 \\
 & 6 & 6 & 4.6 & 0.013 & 0.131 & 0.509 \\
 & 6 & 8 & 5.1 & 0.012 & 0.077 & \textbf{0.439} \\
 & 8 & 4 & 5.1 & 0.012 & 0.165 & 1.043 \\
 & 8 & 6 & 5.7 & 0.017 & 0.171 & 1.020 \\
 & 8 & 8 & 6.2 & 0.016 & 0.082 & 1.091 \\
\cmidrule{2-7}
\multirow{9}{*}{Explicit-Tsf} & 4 & 4 & 4.2 & 0.012 & 0.267 & 0.856 \\
 & 4 & 6 & 5.3 & 0.014 & 0.335 & 1.014 \\
 & 4 & 8 & 6.3 & \textbf{0.011} & 0.314 & 0.970 \\
 & 6 & 4 & 5.3 & 0.025 & 0.389 & 1.128 \\
 & 6 & 6 & 6.3 & \textbf{0.011} & 0.379 & 1.091 \\
 & 6 & 8 & 7.4 & 0.012 & 0.289 & 0.981 \\
 & 8 & 4 & 6.3 & 0.012 & 0.354 & 1.171 \\
 & 8 & 6 & 7.4 & 0.013 & 0.311 & 1.134 \\
 & 8 & 8 & 8.4 & 0.012 & 0.369 & 1.373 \\
\cmidrule{2-7}
\multirow{3}{*}{Implicit} & 4 & - & 2.1 & 0.013 & 0.258 & 0.817 \\
 & 6 & - & 3.2 & \textbf{0.011} & 0.261 & 0.638 \\
 & 8 & - & 4.2 & \textbf{0.011} & 0.287 & 0.661 \\
 \bottomrule
    \end{tabular}
    \caption{Analysis of different design choices for linear regression where implicit and explicit models are tested with different number of parameters or different parameter split between context aggregation and prediction. We conduct experiments on in-distribution generalization as well as out of distribution generalization where either the query is OOD or the underlying latent is.}
    \label{tab:apdx_ablate_1}
\end{table*}

\begin{table*}[]
    \centering
    \footnotesize
    \begin{tabular}{ccccccc}
    \toprule
    & & & & \multicolumn{3}{c}{\textit{Accuracy $(\uparrow)$}} \\
    \cmidrule{5-7}
    \textbf{Model} & \textbf{\# Context Layers} & \textbf{\# Prediction Layers} & \textbf{\# Parameters } (M) & \textbf{IID} & \textbf{OOD Query} & \textbf{OOD Latent} \\
    \midrule
\multirow{3}{*}{Explicit-Known} & 4 & - & 2.1 & 96.381 & 97.779 & 96.288 \\
 & 6 & - & 3.2 & 96.331 & 97.746 & 96.324 \\
 & 8 & - & 4.2 & 96.426 & 97.788 & 96.299 \\
\cmidrule{2-7}
\multirow{9}{*}{Explicit-MLP} & 4 & 4 & 3.0 & 96.297 & 97.646 & 96.150 \\
 & 4 & 6 & 3.6 & 96.176 & 97.410 & 95.989 \\
 & 4 & 8 & 4.1 & 96.338 & 97.772 & 95.964 \\
 & 6 & 4 & 4.1 & 96.289 & 97.679 & 96.125 \\
 & 6 & 6 & 4.6 & 95.885 & 97.154 & 95.803 \\
 & 6 & 8 & 5.1 & 96.167 & 97.503 & 96.039 \\
 & 8 & 4 & 5.1 & 96.194 & 97.496 & 95.761 \\
 & 8 & 6 & 5.7 & 96.158 & 97.508 & 96.224 \\
 & 8 & 8 & 6.2 & 95.915 & 97.006 & 95.758 \\
\cmidrule{2-7}
\multirow{9}{*}{Explicit-Tsf} & 4 & 4 & 4.2 & 96.386 & 97.760 & 96.239 \\
 & 4 & 6 & 5.3 & 96.288 & 97.676 & 96.163 \\
 & 4 & 8 & 6.3 & 96.453 & 97.874 & 96.349 \\
 & 6 & 4 & 5.3 & 96.343 & 97.722 & 96.228 \\
 & 6 & 6 & 6.3 & 96.415 & 97.853 & 96.365 \\
 & 6 & 8 & 7.4 & 96.019 & 97.335 & 96.074 \\
 & 8 & 4 & 6.3 & 96.288 & 97.733 & 96.306 \\
 & 8 & 6 & 7.4 & 95.811 & 96.976 & 95.976 \\
 & 8 & 8 & 8.4 & 96.304 & 97.729 & 96.068 \\
\cmidrule{2-7}
\multirow{3}{*}{Implicit} & 4 & - & 2.1 & 96.525 & 97.996 & 96.149 \\
 & 6 & - & 3.2 & \textbf{96.557} & 98.036 & \textbf{96.440} \\
 & 8 & - & 4.2 & 96.554 & \textbf{98.147} & 96.357 \\
    \bottomrule
    \end{tabular}
    \caption{Analysis of different design choices for linear classification where implicit and explicit models are tested with different number of parameters or different parameter split between context aggregation and prediction. We conduct experiments on in-distribution generalization as well as out of distribution generalization where either the query is OOD or the underlying latent is.}
    \label{tab:apdx_ablate_2}
\end{table*}

\begin{table*}[]
    \centering
    \footnotesize
    \begin{tabular}{cccccc}
    \toprule
    & & & & \multicolumn{2}{c}{\textit{$L_2$ Loss $(\downarrow)$}} \\
    \cmidrule{5-6}
    \textbf{Model} & \textbf{\# Context Layers} & \textbf{\# Prediction Layers} & \textbf{\# Parameters } (M) & \textbf{IID} & \textbf{OOD Query} \\
    \midrule
\multirow{3}{*}{Explicit-Known} & 4 & - & 2.1 & 0.001 & 0.002 \\
 & 6 & - & 3.2 &\textbf{ 0.000} & \textbf{0.001} \\
 & 8 & - & 4.2 & 0.001 & 0.002 \\
\cmidrule{2-6}
\multirow{9}{*}{Explicit-MLP}  & 4 & 4 & 3.0 & 0.001 & 0.295 \\
 & 4 & 6 & 3.6 & 0.001 & 0.254 \\
 & 4 & 8 & 4.1 & 0.001 & 0.208 \\
 & 6 & 4 & 4.1 & 0.001 & 0.298 \\
 & 6 & 6 & 4.6 & 0.001 & 0.307 \\
 & 6 & 8 & 5.1 & 0.001 & 0.229 \\
 & 8 & 4 & 5.1 & 0.001 & 0.399 \\
 & 8 & 6 & 5.7 & 0.002 & 0.266 \\
 & 8 & 8 & 6.2 & 0.001 & 0.222 \\
\cmidrule{2-6}
\multirow{9}{*}{Explicit-Tsf} & 4 & 4 & 4.2 & 0.001 & 0.179 \\
 & 4 & 6 & 5.3 & 0.001 & 0.196 \\
 & 4 & 8 & 6.3 & 0.001 & 0.186 \\
 & 6 & 4 & 5.3 & 0.001 & 0.212 \\
 & 6 & 6 & 6.3 & 0.001 & 0.230 \\
 & 6 & 8 & 7.4 & 0.001 & 0.211 \\
 & 8 & 4 & 6.3 & 0.002 & 0.201 \\
 & 8 & 6 & 7.4 & 0.001 & 0.209 \\
 & 8 & 8 & 8.4 & 0.001 & 0.203 \\
\cmidrule{2-6}
\multirow{3}{*}{Implicit}  & 4 & - & 2.1 & 0.001 & 0.186 \\
 & 6 & - & 3.2 & \textbf{0.000} & 0.143 \\
 & 8 & - & 4.2 & \textbf{0.000} & 0.153 \\
\bottomrule
    \end{tabular}
    \caption{Analysis of different design choices for sinusoid regression where implicit and explicit models are tested with different number of parameters or different parameter split between context aggregation and prediction. We conduct experiments on in-distribution generalization as well as out of distribution generalization.}
    \label{tab:apdx_ablate_3}
\end{table*}

\begin{table*}[]
    \centering
    \footnotesize
    \begin{tabular}{cccccc}
    \toprule
    & & & & \multicolumn{2}{c}{\textit{$L_2$ Loss $(\downarrow)$}} \\
    \cmidrule{5-6}
    \textbf{Model} & \textbf{\# Context Layers} & \textbf{\# Prediction Layers} & \textbf{\# Parameters } (M) & \textbf{IID} & \textbf{OOD Query} \\
    \midrule
\multirow{9}{*}{Explicit-MLP} & 4 & 4 & 3.0 & 0.039 & 0.205 \\
 & 4 & 6 & 3.6 & 0.033 & 0.185 \\
 & 4 & 8 & 4.1 & 0.037 & 0.181 \\
 & 6 & 4 & 4.1 & 0.034 & 0.186 \\
 & 6 & 6 & 4.6 & 0.030 & 0.183 \\
 & 6 & 8 & 5.1 & 0.029 & 0.174 \\
 & 8 & 4 & 5.1 & 0.030 & 0.186 \\
 & 8 & 6 & 5.7 & 0.032 & 0.176 \\
 & 8 & 8 & 6.2 & 0.031 & 0.175 \\
\cmidrule{2-6}
\multirow{9}{*}{Explicit-Tsf} & 4 & 4 & 4.2 & 0.028 & 0.181 \\
 & 4 & 6 & 5.3 & 0.027 & 0.172 \\
 & 4 & 8 & 6.3 & 0.027 & 0.172 \\
 & 6 & 4 & 5.3 & 0.025 & 0.170 \\
 & 6 & 6 & 6.3 & 0.024 & 0.169 \\
 & 6 & 8 & 7.4 & 0.027 & 0.171 \\
 & 8 & 4 & 6.3 & 0.025 & 0.171 \\
 & 8 & 6 & 7.4 & 0.028 & 0.173 \\
 & 8 & 8 & 8.4 & 0.030 & 0.177 \\
\cmidrule{2-6}
\multirow{3}{*}{Implicit} & 4 & - & 2.1 & 0.011 & 0.148 \\
 & 6 & - & 3.2 & \textbf{0.010} & 0.148 \\
 & 8 & - & 4.2 & \textbf{0.010} & \textbf{0.147} \\
\bottomrule
    \end{tabular}
    \caption{Analysis of different design choices for GP regression where implicit and explicit models are tested with different number of parameters or different parameter split between context aggregation and prediction. We conduct experiments on in-distribution generalization as well as out of distribution generalization.}
    \label{tab:apdx_ablate_4}
\end{table*}

\begin{table*}[]
    \centering
    \footnotesize
    \begin{tabular}{cccccc}
    \toprule
    & & & & \multicolumn{2}{c}{\textit{$L_2$ Loss $(\downarrow)$}} \\
    \cmidrule{5-6}
    \textbf{Model} & \textbf{\# Context Layers} & \textbf{\# Prediction Layers} & \textbf{\# Parameters } (M) & \textbf{IID} & \textbf{OOD Query} \\
    \midrule
\multirow{3}{*}{Explicit-Known} & 4 & - & 2.1 & 0.084 & 1.323 \\
 & 6 & - & 3.2 & 0.074 & 1.168 \\
 & 8 & - & 4.2 & 0.073 & \textbf{1.135} \\
\cmidrule{2-6}
\multirow{9}{*}{Explicit-MLP} & 4 & 4 & 3.0 & 0.051 & 2.658 \\
 & 4 & 6 & 3.6 & 0.051 & 2.708 \\
 & 4 & 8 & 4.1 & 0.057 & 2.054 \\
 & 6 & 4 & 4.1 & 0.051 & 2.814 \\
 & 6 & 6 & 4.6 & 0.052 & 2.256 \\
 & 6 & 8 & 5.1 & 0.058 & 2.395 \\
 & 8 & 4 & 5.1 & 0.047 & 2.465 \\
 & 8 & 6 & 5.7 & 0.053 & 2.061 \\
 & 8 & 8 & 6.2 & 0.054 & 1.906 \\
\cmidrule{2-6}
\multirow{9}{*}{Explicit-Tsf} & 4 & 4 & 4.2 & 0.039 & 2.052 \\
 & 4 & 6 & 5.3 & 0.044 & 2.185 \\
 & 4 & 8 & 6.3 & 0.046 & 2.331 \\
 & 6 & 4 & 5.3 & 0.042 & 2.154 \\
 & 6 & 6 & 6.3 & 0.050 & 2.118 \\
 & 6 & 8 & 7.4 & 0.052 & 2.351 \\
 & 8 & 4 & 6.3 & 0.043 & 1.945 \\
 & 8 & 6 & 7.4 & 0.044 & 2.148 \\
 & 8 & 8 & 8.4 & 0.048 & 2.178 \\
\cmidrule{2-6}
\multirow{3}{*}{Implicit} & 4 & - & 2.1 & \textbf{0.024} & 1.443 \\
& 6 & - & 3.2 & \textbf{0.024} & 1.487 \\
& 8 & - & 4.2 & \textbf{0.024} & 1.509 \\
\bottomrule
    \end{tabular}
    \caption{Analysis of different design choices for MLP regression where implicit and explicit models are tested with different number of parameters or different parameter split between context aggregation and prediction. We conduct experiments on in-distribution generalization as well as out of distribution generalization.}
    \label{tab:apdx_ablate_5}
\end{table*}

\begin{table*}[]
    \centering
    \footnotesize
    \begin{tabular}{cccccc}
    \toprule
    & & & & \multicolumn{2}{c}{\textit{Accuracy $(\uparrow)$}} \\
    \cmidrule{5-6}
    \textbf{Model} & \textbf{\# Context Layers} & \textbf{\# Prediction Layers} & \textbf{\# Parameters } (M) & \textbf{IID} & \textbf{OOD} \\
    \midrule
\multirow{3}{*}{Explicit-Known} & 4 & - & 2.1 & 94.426 & 90.753 \\
& 6 & - & 3.2 & 94.633 & 91.086 \\
& 8 & - & 4.2 & 94.492 & 91.443 \\
\cmidrule{2-6}
\multirow{9}{*}{Explicit-MLP} & 4 & 4 & 3.0 & 94.888 & 92.775 \\
 & 4 & 6 & 3.6 & 94.971 & 92.861 \\
 & 4 & 8 & 4.1 & 94.465 & 92.342 \\
 & 6 & 4 & 4.1 & 94.878 & 92.807 \\
 & 6 & 6 & 4.6 & 94.933 & 93.024 \\
 & 6 & 8 & 5.1 & 94.851 & 93.097 \\
 & 8 & 4 & 5.1 & 94.818 & 93.036 \\
 & 8 & 6 & 5.7 & 94.822 & 93.024 \\
 & 8 & 8 & 6.2 & 94.878 & 92.985 \\
\cmidrule{2-6}
\multirow{9}{*}{Explicit-Tsf} & 4 & 4 & 4.2 & 95.074 & 92.990 \\
 & 4 & 6 & 5.3 & 95.060 & 92.988 \\
 & 4 & 8 & 6.3 & 95.157 & 93.190 \\
 & 6 & 4 & 5.3 & 94.989 & 93.010 \\
 & 6 & 6 & 6.3 & 95.060 & 93.275 \\
 & 6 & 8 & 7.4 & 95.078 & 93.119 \\
 & 8 & 4 & 6.3 & 95.010 & 93.133 \\
 & 8 & 6 & 7.4 & 95.058 & 93.121 \\
 & 8 & 8 & 8.4 & 94.969 & 92.954 \\
\cmidrule{2-6}
\multirow{3}{*}{Implicit} & 4 & - & 2.1 & 95.285 & \textbf{93.394} \\
& 6 & - & 3.2 & \textbf{95.325} & 93.336 \\
& 8 & - & 4.2 & 95.324 & 93.351 \\
\bottomrule
    \end{tabular}
    \caption{Analysis of different design choices for MLP classification where implicit and explicit models are tested with different number of parameters or different parameter split between context aggregation and prediction. We conduct experiments on in-distribution generalization as well as out of distribution generalization.}
    \label{tab:apdx_ablate_6}
\end{table*}

\begin{table*}[]
    \centering
    \footnotesize
    \begin{tabular}{cccccc}
    \toprule
    & & & & \multicolumn{2}{c}{\textit{Accuracy $(\uparrow)$}} \\
    \cmidrule{5-6}
    \textbf{Model} & \textbf{\# Context Layers} & \textbf{\# Prediction Layers} & \textbf{\# Parameters } (M) & \textbf{IID} & \textbf{OOD} \\
    \midrule
\multirow{3}{*}{Explicit-Known} & 4 & - & 2.1 & 99.864 & 99.844 \\
& 6 & - & 3.2 & \textbf{99.876} & \textbf{99.875} \\
& 8 & - & 4.2 & 99.545 & 99.539 \\
\cmidrule{2-6}
\multirow{3}{*}{Explicit-MLP} & 4 & 4 & 3.4 & 97.929 & 97.894 \\
 & 4 & 6 & 4.0 & 97.551 & 97.532 \\
 & 4 & 8 & 4.5 & 97.804 & 97.802 \\
 & 6 & 4 & 4.5 & 97.394 & 97.383 \\
 & 6 & 6 & 5.0 & 97.524 & 97.517 \\
 & 6 & 8 & 5.5 & 98.527 & 98.514 \\
 & 8 & 4 & 5.5 & 29.883 & 29.587 \\
 & 8 & 6 & 6.1 & 98.240 & 98.216 \\
 & 8 & 8 & 6.6 & 96.726 & 96.677 \\
\cmidrule{2-6}
\multirow{9}{*}{Explicit-Tsf} & 4 & 4 & 4.2 & 98.522 & 98.513 \\
& 4 & 6 & 5.3 & 98.467 & 98.455 \\
 & 4 & 8 & 6.3 & 98.006 & 98.005 \\
 & 6 & 4 & 5.3 & 98.494 & 98.459 \\
 & 6 & 6 & 6.3 & 97.930 & 97.874 \\
 & 6 & 8 & 7.4 & 98.566 & 98.591 \\
 & 8 & 4 & 6.3 & 98.280 & 98.260 \\
 & 8 & 6 & 7.4 & 97.750 & 97.744 \\
 & 8 & 8 & 8.5 & 98.468 & 98.456 \\
\cmidrule{2-6}
\multirow{3}{*}{Implicit} & 4 & - & 2.1 & 98.359 & 98.334 \\
 & 6 & - & 3.2 & 98.332 & 98.309 \\
 & 8 & - & 4.2 & 97.164 & 97.176 \\
\bottomrule
    \end{tabular}
    \caption{Analysis of different design choices for RAVEN's progressive matrices where implicit and explicit models are tested with different number of parameters or different parameter split between context aggregation and prediction. We conduct experiments on in-distribution generalization as well as compositional out of distribution generalization.}
    \label{tab:apdx_ablate_7}
\end{table*}

\begin{table*}[]
    \centering
    \footnotesize
    \begin{tabular}{ccccc}
    \toprule
    & & & & \multicolumn{1}{c}{\textit{$R^2$ $(\uparrow)$}} \\
    \cmidrule{5-5}
    \textbf{Model} & \textbf{\# Context Layers} & \textbf{\# Prediction Layers} & \textbf{\# Parameters } (M) & \textbf{Performance} \\
    \midrule
\multirow{9}{*}{Explicit-MLP} & 4 & 4 & 9.5 & 0.815 \\
& 4 & 6 & 10.0 & \textbf{0.827} \\
& 4 & 8 & 10.5 & 0.814 \\
& 6 & 4 & 10.5 & 0.814 \\
& 6 & 6 & 11.0 & 0.818 \\
& 6 & 8 & 11.6 & 0.819 \\
& 8 & 4 & 11.6 & 0.828 \\
& 8 & 6 & 12.1 & 0.820 \\
& 8 & 8 & 12.6 & 0.817 \\
\cmidrule{2-5}
\multirow{9}{*}{Explicit-Tsf} & 4 & 4 & 8.1 & 0.829 \\
& 4 & 6 & 9.1 & 0.824 \\
& 4 & 8 & 10.2 & 0.823 \\
& 6 & 4 & 9.1 & 0.808 \\
& 6 & 6 & 10.2 & 0.812 \\
& 6 & 8 & 11.3 & 0.812 \\
& 8 & 4 & 10.2 & 0.810 \\
& 8 & 6 & 11.3 & 0.818 \\
& 8 & 8 & 12.3 & 0.821 \\
\cmidrule{2-5}
\multirow{3}{*}{Implicit} & 4 & - & 4.7 & 0.819 \\
& 6 & - & 5.7 & 0.823 \\
& 8 & - & 6.8 & 0.817 \\   
\bottomrule
\end{tabular}
    \caption{Analysis of different design choices for gene targeting experiments where implicit and explicit models are tested with different number of parameters or different parameter split between context aggregation and prediction. We conduct experiments on compositional out of distribution generalization.}
    \label{tab:apdx_ablate_8}
\end{table*}
\end{document}